\documentclass[10pt,twocolumn,letterpaper]{article}
\usepackage{wacv}
\usepackage{times}
\usepackage{epsfig}
\usepackage{graphicx}
\usepackage{amsmath}
\usepackage{amssymb}
\usepackage{booktabs}
\usepackage{enumitem}
\usepackage{bbm}
\usepackage{multirow}
\usepackage{colortbl}
\usepackage{booktabs}       
\usepackage{cuted}
\usepackage{capt-of}
\usepackage{verbatim}
\usepackage{tabularx}
\usepackage[accsupp]{axessibility}

\definecolor{color_1}{RGB}{255,0,128}
\definecolor{color_2}{RGB}{0,128,128}
\definecolor{color_3}{RGB}{0,128,0}
\definecolor{color_4}{RGB}{128,0,0}
\definecolor{color_5}{RGB}{128,0,128}

\newcommand{\ShortName}[0]{\textsc{Seg\&Struct}}

\def\wacvPaperID{1352}
\wacvalgorithmstrack 
\wacvfinalcopy 
\ifwacvfinal
\usepackage[breaklinks=true,bookmarks=false]{hyperref}
\else
\usepackage[pagebackref=true,breaklinks=true,colorlinks,bookmarks=false]{hyperref}
\fi
\begin{document}

\title{\ShortName{}: The Interplay Between Part Segmentation and Structure Inference for 3D Shape Parsing}

\author{
    Jeonghyun Kim$^{1}\quad$ 
    Kaichun Mo$^{2}\quad$ 
    Minhyuk Sung$^{1\dag}$\quad 
    Woontack Woo$^{1\dag}$
    \\
    $^{1}$KAIST \quad 
    $^{2}$Stanford University
}

\twocolumn[
{%
    \renewcommand\twocolumn[1][]{#1}%
    \maketitle
    \thispagestyle{empty}
    \begin{center}
    \centering
        \includegraphics[width=\textwidth]{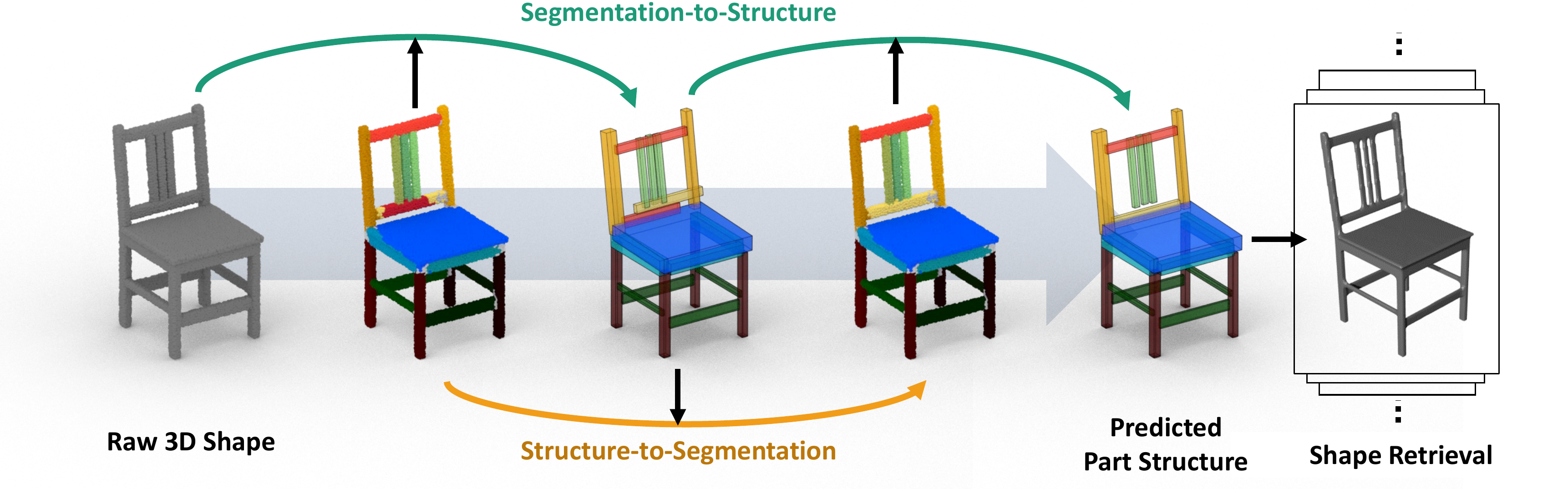}
        \captionof{figure}{\textbf{Overivew.} 
        We introduce \ShortName{}, a novel framework for the interplay between part segmentation and structure inference. 
        In the forward path (green), our framework parses a raw 3D shape into part segments and predicts a part structure driven by an established \emph{point-to-part associations}. In the backward path (yellow), the predicted structure is leveraged for segmentation refinement to relax the confusion of part boundaries. Through this interplay, we can achieve a more accurate part structure and improved segmentation. 
        We also showcase that the predicted structure further can be used for shape retrieval by measuring a structural similarity between two shapes.
        \label{fig:teaser}}
    \end{center}%
}
]

\def\thefootnote{\dag}
\footnotetext{ Co-corresponding Authors.}
\def\thefootnote{\arabic{footnote}}

\begin{abstract}
We propose \ShortName{}, a supervised learning framework leveraging the interplay between part segmentation and structure inference and demonstrating their synergy in an integrated framework. 
Both part segmentation and structure inference have been extensively studied in the recent deep learning literature, while the supervisions used for each task have not been fully exploited to assist the other task. 
Namely, structure inference has been typically conducted with an autoencoder that does not leverage the point-to-part associations. 
Also, segmentation has been mostly performed without structural priors that tell the plausibility of the output segments. 
We present how these two tasks can be best combined while fully utilizing supervision to improve performance. 
Our framework first decomposes a raw input shape into part segments using an off-the-shelf algorithm, whose outputs are then mapped to nodes in a part hierarchy, establishing point-to-part associations.
Following this, ours predicts the structural information, e.g., part bounding boxes and part relationships. 
Lastly, the segmentation is rectified by examining the confusion of part boundaries using the structure-based part features. 
Our experimental results based on the StructureNet and PartNet demonstrate that the interplay between two tasks results in remarkable improvements in both tasks: 27.91\% in structure inference and 0.5\% in segmentation. 
\end{abstract}

\section{Introduction}
The importance of compositional understanding of 3D shapes has been reiterated for a long time in computer graphics and computer vision.
Until recently, a large body of work has investigated how the part structure of 3D shapes can be utilized in various applications including 
object recognition~\cite{10.1145/2366145.2366157, 10.1145/2366145.2366156}, shape completion~\cite{sung2015data, 10.1145/3130800.3130821}, shape editing~\cite{li2017grass, 10.1145/3355089.3356527, mo2020structedit}, 
deformation~\cite{10.1145/1576246.1531341, 10.1145/2185520.2185574}, functional attributes analysis~\cite{10.1145/2461912.2461924, mo2021where2act, huang2021multibodysync, 10.1145/3414685.3417783}, shape retrieval~\cite{izadinia2017im2cad,avetisyan2019scan2cad, dahnert2019joint,uy2021joint}, and so on.

Due to the vast range of applications, there also have been lots of previous studies about learning representations of the part structure of 3d shapes, such as GRASS~\cite{li2017grass} and StructureNet~\cite{10.1145/3355089.3356527}.
StructureNet particularly offers highly curated information about the part structure including the part abstraction (e.g., part bounding boxes) and the relationships across the parts (e.g., symmetry across sibling parts).
A straightforward idea leveraging such supervision for the part structure inference from a raw 3D shape is to build a simple encoder-decoder network, whose encoder takes the raw 3D shape and outputs a latent code, and the decoder takes the code and produces the part structure. 
The limitation of such an approach is, however, that the neural network does not exploit the supervision of association between the areas on the raw 3D shape and the semantic parts. 


3D segmentation is another direction of learning the compositional structure from 3D shapes but focusing on decomposition. 
It also has been extensively studied in the deep learning literature, particularly to segment object instances from a scene~\cite{jiang2020pointgroup, vu2022softgroup}. 
While recent segmentation techniques provide accurate results in many cases, the methods are still limited to learning the point-to-semantic part relationships without considering the global structure resulting from the segmentation. 
The global structural priors can assist in pruning such noises affecting the structure, while such as idea has not been explored in recent deep-learning-based approaches.

To overcome the limitations in both tasks, we propose~\ShortName{}, a framework incorporating the interplay between them while fully exploiting the supervision of both part segmentation and point-to-part associations.
Our pipeline is divided into two tasks: 1) a \emph{Segmentation-to-structure inference} and 2) \emph{Structure-to-segmentation refinement}. 
To infer a part structure from a raw 3d shape, we first extract part segments from the input geometry and then map them to nodes in a part hierarchy.
This two-step approach establishing a correspondence between the part segments and nodes in the part hierarchy greatly helps the predicted structure resemble the input geometry. 
Subsequently, we take an additional step that uses the part structures to draw the candidates of incorrectly split parts and prune out the candidates assisted by the structure-aware features from the previous step.
This reverse process and its rectification of the noisy segments close the loop of the pipeline and demonstrate the synergy between the two tasks. 

Our experimental results based on the StructureNet dataset demonstrate our method outperforms the baselines using an encoder-decoder network, with significant margins of 27.91\%. Our results also show that the segmentation can be better refined with the predicted structural information by 0.5\% in the PartNet dataset. 

To summarize, our contributions are following:

\begin{itemize} [leftmargin=*,itemsep=2pt,parsep=2pt]
\item We propose a supervised learning framework leveraging the interplay between part segmentation and structure inference and demonstrate its synergy in improving performance.
\item We introduce a structure prediction module that takes advantage of a part segmentation network exploiting the supervision of point-to-part associations. 
\item We also introduce a part segmentation refinement module that learns the confusion of segmentation boundary from the predicted structure.
\item Our experimental results demonstrate significant outperformance of our integrated framework compared with previous methods. 
\end{itemize}



\section{Related Work}
Our work is primarily related to the two threads of research study on segmenting 3D shapes into parts and inferring the structure of 3D shapes.
While both directions have been extensively explored in the literature, 
we propose to tackle the two tasks jointly and leverage their synergy to boost the performance for both tasks in this paper.
We briefly review the two fields of study below.

\subsection{Part Segmentation on 3D Shapes} 

Given a raw input 3D geometry, segmenting it into 3D parts is a long-standing important yet challenging research problem in 3D computer graphics and vision.
Prior to the popularity of data-driven methods, early works have investigated various optimization based techniques for segmenting 3D mesh inputs into parts~\cite{10.1145/1409060.1409098, 10.1145/2611811, liu2009part, 10.1145/2010324.1964928,wang2012active, attene2006mesh,shamir2008survey,chen2009benchmark}.
Recently, driven by the advancement of modern machine learning techniques and the availability of large-scale data sets~\cite{chang2015shapenet,yi2016scalable,mo2019partnet,yu2019partnet},
researchers have switched gears to work on data-driven solutions.
While many works proposed learning based approaches to perform 3D semantic part segmentation -- assigning semantic labels (\textit{e.g.}, chair back, seat and base) to each point or face over the input geometry, such as~\cite{kalogerakis2010learning,wang2013projective,qi2016pointnet,Kalogerakis:2017:ShapePFCN,yi2017syncspeccnn}, more related to us are the previous studies on 3D instance part segmentation where all different part instances (\textit{e.g.}, the four chair legs) are separated, \textit{e.g.}, ~\cite{mo2019partnet,luo2020learning,zhang2021point}.
There are also many works mostly focusing on 3D scene instance segmentation but also demonstrating good results on segmenting 3D shapes into parts, including~\cite{wang2018sgpn,wang2019associatively,han2020occuseg,jiang2020pointgroup}.
Researchers have been pushing state-of-the-art for 3D shape part segmentation using such learning-based methods driven by large-scale training data with part labels.
While these works show impressive results, they are limited to decomposing the shape into parts, not connecting the parts to semantic and structural priors such as the part names and the relationships across the parts (e.g., parent-child, symmetry, etc).

In the vast literature on 3D shape part segmentation, only a few past works have explicitly exploited the structural information of 3D shapes.
For example, researchers have attempted to leverage part templates~\cite{Kim13,ganapathi2018parsing}, part hierarchy~\cite{Yi17,wang2018learning,yu2019partnet}, and shape grammar~\cite{jonesneurally} to capture the rich part relationships and constraints.
While these works have demonstrated that the 3D part segmentation tasks benefit from estimating the shape structure, they have not explored if predicting 3D shape parts can inversely suggest better shape structural predictions. 
Our work studies and confirms the synergy between the two tasks.

\subsection{3D Shape Structure Inference}
3D objects are often highly structured in their geometry, parts, and rich part relationships.
For example, a physically stable chair is often governed by a set of rules specifying some strong relationships and constraints among the shape parts, \textit{e.g.}, the four legs are distributed symmetrically and of the same length.
Therefore, researchers have been investigating various approaches to infer the 3D shape structure from raw geometry input.
Previous works have proposed different ways to represent the structure of 3D shapes, such as part-based templates~\cite{10.1145/2010324.1964928,Kim13,ganapathi2018parsing}, part-level or shape-level symmetry~\cite{mitra2006partial,sung2015data}, shape programs~\cite{muller2006procedural,Tian2019LearningTI,jones2020shapeassembly}, shape grammars~\cite{Chaudhuri2011ProbabilisticRF,kalogerakis2012probabilistic}, \textit{etc.}.
After estimating the 3D shape structure, these works can then leverage such information to perform diverse downstream tasks, such as shape generation~\cite{Tian2019LearningTI,jones2020shapeassembly}, editing~\cite{kalogerakis2012probabilistic,ganapathi2018parsing} and completion~\cite{sung2015data}.

Recent works have explored designing learning models that represent and model the shape structure as part hierarchies or graphs~\cite{li2017grass,Yi17,zhu2018scores,wu2019sagnet,10.1145/3355089.3356527,10.1145/3355089.3356488,yang2020dsg}.
Among these works, GRASS~\cite{li2017grass} is the pioneering work designing a novel learning framework to encode and decode binary part hierarchies, 
while StructureNet~\cite{10.1145/3355089.3356527} further extends the system to handle \textit{n}-ary hierarchical part trees and the rich part relationships.
Follow-up works have tried to predict the hierarchical shape structure from a single input image of the 3D shape~\cite{niu2018im2struct,Paschalidou2020CVPR} or a 3D input point cloud~\cite{yu2019partnet,mo2020pt2pc}.
Given the good performance in modeling 3D shape structure, in this paper, we adopt the hierarchical and relational structural shape representation introduced in StructureNet~\cite{10.1145/3355089.3356527} and focus on investigating the interplay between the two tasks of structure prediction and part segmentation.

\section{The Interplay-based Framework}
\subsection{Overview}
Our goal is to build a synergy between the part segmentation and the structure inference via the bi-directional interplay between two tasks in an integrated framework.
To this end, we propose two separate networks: a \emph{Segmentation-to-Structure Inference} network (Sec~\ref{Forward Path}) and a \emph{Structure-to-Segmentation} network (Sec~\ref{Backward Path}). 
The first network predicts a part structure from a given raw input geometry by exploiting an association between part regions and part nodes in the structure.
Afterward, the second network relaxes the confusion of part segment boundaries by leveraging the structure-aware features derived from the earlier stage.

\textbf{Notations.} We denote the raw 3D shape as a point cloud $A=\{a_1,...a_N\}$ for each point $a\in\mathbb{R}^3$, and a part segmentation output from $A$ as $B=\{b_l\}_{l \in L}$, where $b_l$ is $l$-th part segment which contains a set of points in the part region~$X_l$ and its semantic label~$y_l$, and $L$ is the number of part segments.
For the part structure, we use a tree representation $S=(P, \textbf{H}, \textbf{R})$ defined in StructureNet~\cite{10.1145/3355089.3356527}. 
Here, a set of part nodes is represented by $P$, and relationships for the hierarchical connection and part relations are denoted as \textbf{H} and \textbf{R}, respectively. 
For $M$ number of parts $P=\{m_1,...,m_M\}$, a part node $m_i$ contains a 128-dimensional feature vector and an one-hot encoded semantic label $y_i$ as $m_i=(\textbf{x}_i,y_i)$.
At the end of the structure inference stage, the network predicts a part geometry represented in oriented bounding box parameters $\theta_i=(\textbf{t}_i, \textbf{s}_i, q_i)$ based on the part feature $\textbf{x}_i$. 
The box parameters $\theta_i$ include a translation vector $\textbf{t}_i \in \mathbb{R}^3$, a scaling vector $\textbf{s}_i \in \mathbb{R}^3$, and an orientation in unit quaternion $q_i\in\mathbb{H}$.
To describe a part relation between two nodes, we use a edge $(m_i,m_j,\tau)$ for a part relation type~$\tau\in\mathcal{T}$, where $\mathcal{T}$ is a pre-defined set of part relations, e.g. symmetry and adjacency. 

 \begin{figure*}
     \centering
     \includegraphics[width=0.95\textwidth]{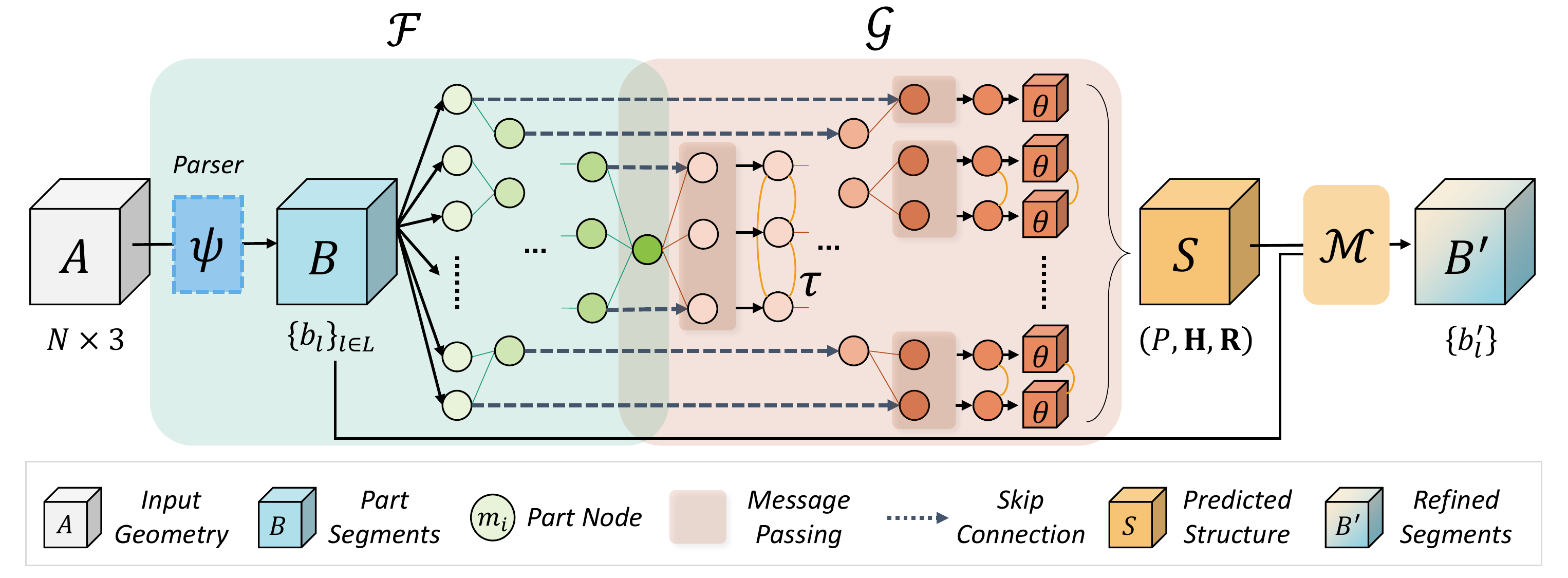}
     \caption{\textbf{Overall Framework Architecture.} 
     First, the structure encoder $\mathcal{F}$ decomposes an input geometry $A$ into part segments $B=\{b_l\}$ using the parser $\psi$. 
     By treating these segments as the leaf nodes, $\mathcal{F}$ constructs a tree hierarchy in a bottom-up manner and aggregates the features of sibling nodes to create a feature vector of their parent node. 
     The following structure decoder $\mathcal{G}$ propagates the root feature throughout the hierarchy in a top-down manner and aggregates the features of parent and sibling nodes indicated by the skip connection. 
     Further, the refinement network $\mathcal{M}$ relaxes the confusion of part boundaries in the segmentation by leveraging the predicted structure.} 
     \label{fig:overview}
 \end{figure*}

\subsection{Segmentation-to-Structure Inference} \label{Forward Path}
We first propose a parsing-based structure encoder~$\mathcal{F}$ that exploits the \emph{segmentation prior}.
To infer part structure from a raw 3D shape, one can naively utilize a simple autoencoder network, which encodes the input ~$A\in\mathbb{R}^{N\times3}$ into a latent code and decodes it into a part structure~$S$, adopting a decoder similar to StructureNet~\cite{10.1145/3355089.3356527}. 
However, this approach hardly yields an accurate output.
Since the information of 3D shape is just aggregated into a single latent code, the network does not see which point in the input shape belongs to which part in the structure.

To address this, we take a two-step approach that parses the input geometry~$A$ into a set of part segments $B$ using a parser backbone~$\psi$ and then construct a structure hierarchy~$\textbf{H}$.
For $\psi$, we utilize PointGroup \cite{jiang2020pointgroup}, an off-the-shelf algorithm for 3d scene instance segmentation, by treating part instances in a single object as object instances in an indoor scene.
To build the structure hierarchy, we treat these parsed segments $B$ as leaf nodes and group them in a bottom-up manner recursively until we get a root node, according to the rule defined in StructureNet~\cite{10.1145/3355089.3356527}, which is based on the part labels.
For example, a set of part nodes with a semantic label named \textit{leg} has to be grouped together as sibling nodes under a \textit{base} node.
While the hierarchy is built starting from leaf nodes, the network also encodes each part geometry into a feature vector~$\textbf{x}_i\in \mathbb{R}^{128}$ and aggregates the vectors of sibling nodes to produce a feature of their parents in the same dimensionality. 
At the end of the encoding step, we get the root feature vector $\textbf{x}^{root}\in\mathbb{R}^{128}$.
Finally, we can establish a correspondence between part regions in input geometry and part nodes in structure hierarchy, which largely helps the later network to associate the given shape and the part structure. 

Next, we introduce a structure decoder $\mathcal{G}$, which recursively propagates the root feature $\textbf{x}^{root}$ to leaf nodes, and predicts the part bounding boxes $\{\theta_i\}$ and the part relations $\{(m_i,m_j,\tau)\}$ between two nodes under the same parent node. 
For $\mathcal{G}$, we adapt a similar decoder network proposed in StructureNet~\cite{10.1145/1409060.1409098}, which uses the message passing network across the part features in the hierarchy.
Through message passing, the network updates the part features to contain a much broader context across the hierarchy. We denote this updated part feature as $\textbf{x}'_i\in\mathbbm{R}^{128}$.
However, it is not feasible to directly use the decoder from StructureNet, which produces the whole part structures by recursively decoding a single latent vector at the root node.
When the input is given in the form of the raw geometry, it suffers to predict the existence of part nodes due to the domain gap between the geometry input and the structure output.

To tackle this, our decoder takes advantage of an \emph{explicit} guidance given by the established correspondence between the part regions and part nodes in the previous stage. 
The previous structure construction step enables us to represent each part segment in the input shape as the hierarchical representation. 
Based on this, we can associate given part regions to the corresponding part nodes using \emph{skip connection} (Figure~\ref{fig:overview}).  
This association gives us strong supervision for the structure decoding step and enables the network to know the exact part region to which the part node is related.
As a result, our network becomes less dependent on the implicit latent code and infers the part structure resembling the input geometry.
We will discuss how this affects the performance of structure inference in the experiments (Sec~\ref{exp_forward}).

\subsection{Structure-to-Segmentation Refinement} \label{Backward Path}
In this section, we introduce a segmentation refinement network $\mathcal{M}$ on top of the structure inference. 
By predicting the merge operation at the confusion of part regions, our network relaxes noisy regions from the first segmentation output.
We found that this task largely depends on the structural context since the local information does not suffice to decide which part has to be merged to other one. 
To this end, we exploit the features from earlier inference stage, which play a critical role for merge prediction. 

\begin{figure}
    \centering
    \includegraphics[width=0.48\textwidth]{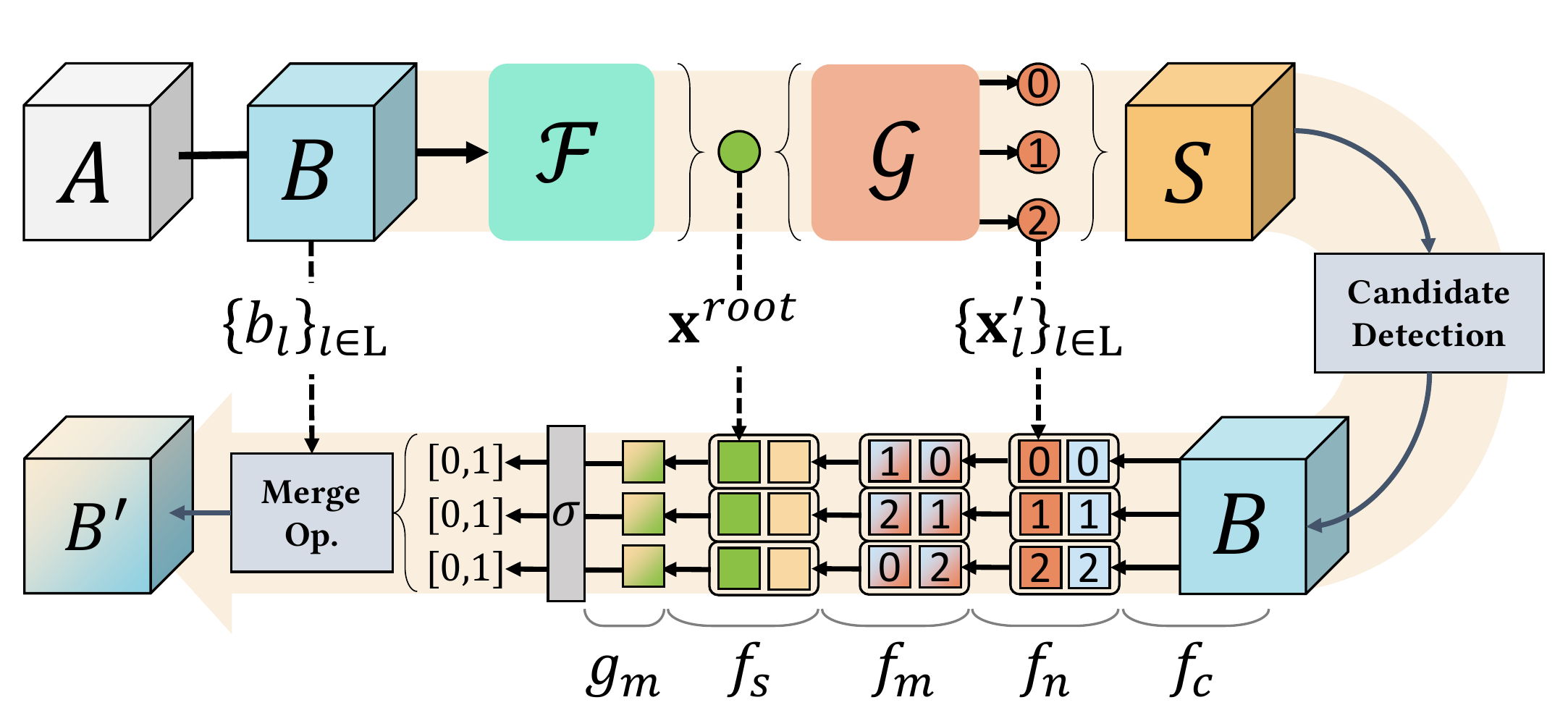}
    \caption{\textbf{Structure-induced Segmentation Refinement.} 
    Our network first sample the part segments to be merged in predicted structure $S$ through candidate detection.
    The number in the box from part segments $B$ indicates the index of each part region.
    The order of indices inside the boxes grouped together means the direction of merge, e.g., 1-0 means the 1-th part segment can be merged to the 0-th part.
    Utilizing a series of features from $S$, refinement network further predicts merge scores for candidate pairs. 
    }
    \label{fig:method for Structure-induced Segmentation Refinement}
\end{figure}

To detect the confusion of part boundaries, we apply a simple heuristic using \emph{Intersection over Union}~(IoU). 
An IoU is computed using predicted part boxes and treated as \emph{conflict score}.
We filter out the candidates with a larger score than a threshold, i.e. 0.09. 
Note that if there are multiple confusions for one node, we take only one candidate with the largest score.
This means we only consider one-directional merge cases where the merge operation is order-variant, e.g. a merge candidate for one node does not have to be vice-versa, as shown in Figure~\ref{fig:method for Structure-induced Segmentation Refinement}.
The valid candidate pairs are assigned to $\textbf{C}$, a $M \times M$ binary matrix having each row as an index of the part node to be merged and each column as an index of the target node. 
For example, $\textbf{C}_{(i,j)}=1$ describes $i$-th node have a chance to be merged to $j$-th node.   

To predict merge operations across the nodes in $\textbf{C}$, we first compute a candidate feature vector $\textbf{c}_i\in\mathbb{R}^{256}$ for each candidate part by encoding the part segment $b_i=(X_i;y_i)$ using a candidate feature encoder $f_c$.
For $f_c$, we use a vanilla PointNet~\cite{qi2016pointnet} architecture where we treat a part segment as a single point. 
\begin{equation}
\label{eq:candidate feature encoding}
   \textbf{c}_{i} = f_c([X_i;y_i])
\end{equation}
However, the feature $\textbf{c}_{i}$ contains only fragmentary information of the local part region.
Therefore, we opt to update the candidate feature with its corresponding part node feature $\textbf{x}'_i$ from the inference stage.
Using a single linear layer $f_n$, we aggregate this part-wise structural information to $\textbf{c}_{i}$ producing an updated candidate feature as $\tilde{\textbf{c}}_{i} = f_n([\textbf{c}_{i};\textbf{x}_{i}])$.

After updating the candidate features, we compute a merge feature, an order-variant feature vector to predict merge operation from the candidate node to their defined target node.
We denote the merge feature as a 256-dimensional vector $\textbf{m}_{ij}\in\mathbb{R}^{256}$ and merge feature encoder as $f_{m}$, which uses a single linear layer.
\begin{equation}
\label{eq:candidate detection}
    \textbf{m}_{ij} = f_{m}([\tilde{\textbf{c}}_{i};\tilde{\textbf{c}}_{j}])
\end{equation}
Similar to the candidate feature case, we update $\textbf{m}_{ij}$ to assist more comprehensive structural context to it from the predicted structure.
To this end, we brought the root node feature $\textbf{x}^{root}$ and treat it as a \emph{structure code} and aggregate it to all merge features.
We use an another single linear layer encoder $f_{s}$, which outputs the updated merge feature $\tilde{\textbf{m}}_{ij} = f_{s}([ \textbf{m}_{ij};\textbf{x}^{root}])$.

After the series of the concatenations and feature encoding, we finally predict the probability score of each merge operation~$\in[0,1]$ similar to edge prediction: 
\begin{equation}
\label{eq:merge prediction}
    p^{merge}_{(m_i, m_j)} = \sigma(g_m(\tilde{\textbf{m}}_{ij}))
\end{equation}
where $g_m$ decodes the edge feature into logits, $\sigma$ and $p$ are the sigmoid and probability function, respectively.
For the pairs with merge scores larger than the threshold (i.e., 0.7), we perform merge operation to update the first part segmentation by attaching the source part segments to their targets.

\subsection{Training and Loss}
Our goal is to train a category-specific framework that integrates two networks: structure inference and segmentation refinement.
We train two networks separately and use the freezed part segmentation backbone $\psi$, which is pretrained in advance.

First, the parameters for the encoder $\mathcal{F}$ and the decoder $\mathcal{G}$ are supervised both at the structure decoding step through backpropagation. 
Our loss design is mostly brought from StructureNet~\cite{10.1145/3355089.3356527}, which computes geometry loss for part bounding boxes, edge prediction loss for part relations, and structure consistency loss to make the relations at parent nodes transfer to their siblings. 
For more detail, we refer the readers to the original paper and our supplementary.

The training for our refinement network $\mathcal{M}$ is considered as the traditional binary classification problem.
However, we face the imbalanced data distribution problem having the majority of merge operation label \emph{True Negative}, which means most of the candidate pairs should not be merged due to the fairly good quality of the previous part segmentation output.
To address this, we use a \emph{Focal Loss}~\cite{lin2017focal}, a modified version of binary cross entropy loss to handle the data imbalance problem by setting a bigger weight for a label with a sparse number of samples.
Please refer to the supplementary for more detail on training.


\begin{figure*}
    \centering
    \includegraphics[width=\textwidth]{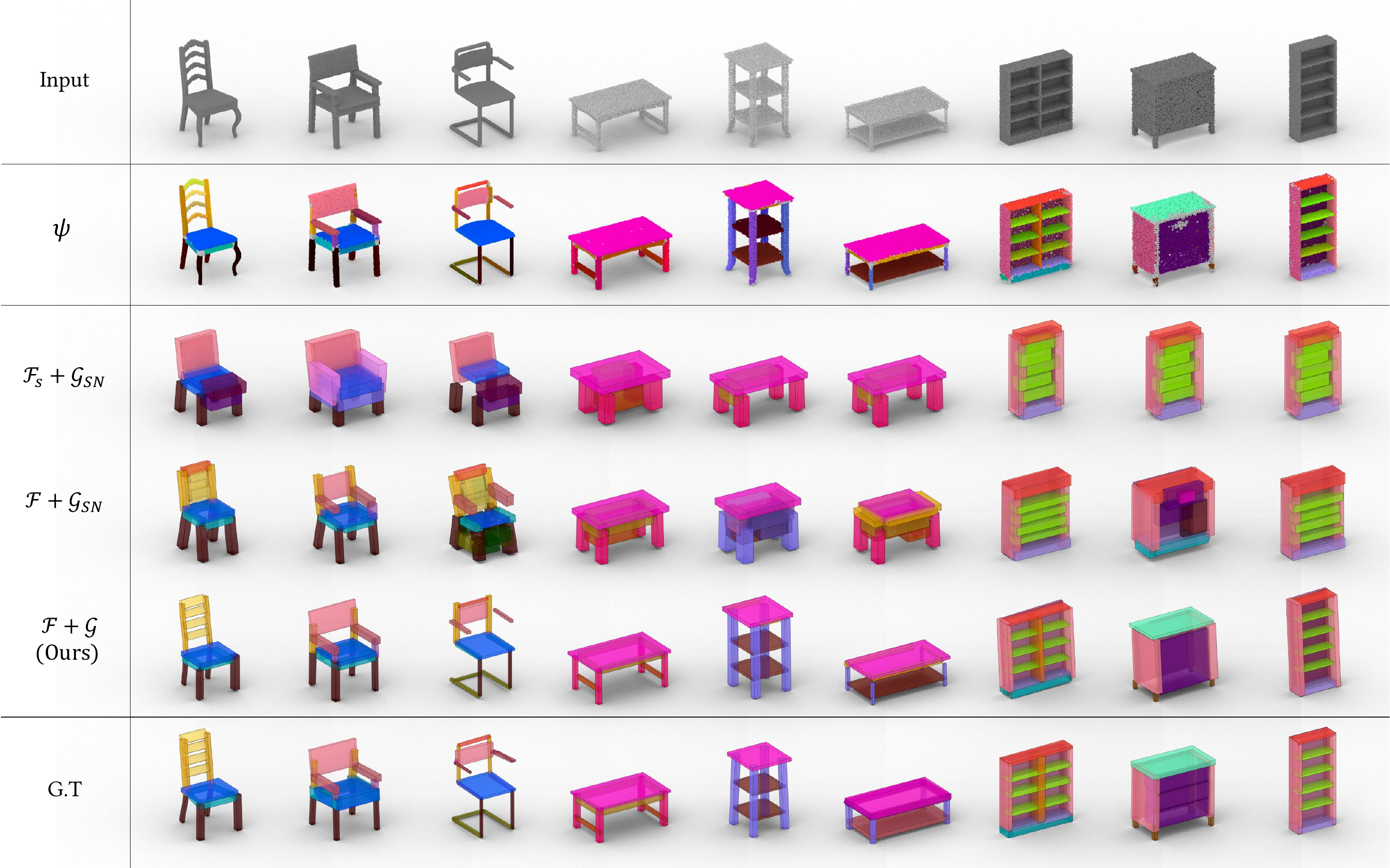}
    \caption{\textbf{Qualitative Comparison on Structure Inference.} The top row describes the input 3D shape and the bottom row describes ground-truth. In the second row, the part segmentation outputs from $\psi$ are shown. Compared to other baselines, ours~(bottom row) achieves the most accurate part structures also capturing a more diverse set of part structures.} 
    \label{fig:results_forward}
\end{figure*}

\section{Experimental Results}
In this section, we demonstrate our experimental results for two main tasks and one application: structure inference, segmentation refinement, and structure-aware shape retrieval.
For more results including ablation study and discussion on failure cases, please refer to our supplementary.

\textbf{Data Preparation.}
We prepared two kinds of datasets to test our method: PartNet~\cite{mo2019partnet} and StructureNet~\cite{10.1145/3355089.3356527}.
ParNet provides point cloud data sampled on surfaces of 3D mesh from ShapeNet~\cite{chang2015shapenet} and its corresponding semantic-instance part annotation.
We use PartNet to train our segmentation backbone $\psi$ and evaluate the performance of segmentation refinement.
StructureNet is built upon PartNet with an additional annotation on the structure hierarchy, part bounding boxes, and part relations. 
Same as StructureNet, we set  a maximum number of parts in a subset of the tree as 10 and four types of part relations, i.e. translational, rotational, reflective symmetry, and adjacency.
To test our method, we pick three shape categories from StructureNet, which have diverse and complex structures compared to other shapes: \textit{chair}, \textit{table}, and \textit{storage furniture}.
Since our framework learns from two datasets at the same time, we filter the invalid shape missing one of the annotations from them.  
In total, the remaining shapes for chair, table, and storage furniture are 3522, 1802, and 932, respectively.
We split these samples into the train and test set following PartNet.

\subsection{Evaluation on Structure Inference}\label{exp_forward}
\textbf{Baselines.}
Since there are no directly comparable methods in previous studies, we build two encoder-decoder baselines.
The first one~($\mathcal{F}_s+\mathcal{G}_{SN}$) encodes the input into a single feature vector using a shape encoder $\mathcal{F}_s$ without any segmentation prior and decodes it to predict the whole part structure using a structure decoder $\mathcal{G}_{SN}$.
We use PointNet++ \cite{qi2017pointnetplusplus} for $\mathcal{F}_s$.
and use the same decoder from StructureNet~\cite{10.1145/3355089.3356527} for $\mathcal{G}_{SN}$.
Meanwhile, the second baseline~($\mathcal{F}+\mathcal{G}_{SN}$) encodes the extracted part segments into a root feature in the hierarchy using our encoder $\mathcal{F}$ and decodes the vector sharing the same decoder $\mathcal{G}_{SN}$, not exploiting the point-to-part association.
Both baselines expect the latent vector to contain all the information for the part structure without using one or any priors used in our method. 

\textbf{Metrics.}
We evaluate our method using two metrics: 1) part prediction accuracy and 2) edge prediction error.
The part prediction accuracy measures how accurately are part bounding boxes predicted. 
We calculate this accuracy using \emph{Average Precision (AP)}, which is widely used in the object detection problem~\cite{song2016deep}.
Since the part semantics are given from segmentation backbone~$\psi$, we use a class-agnostic AP with an IoU threshold 0.25.
Note that the correspondence between the prediction and target structure is established only for the leaf nodes.
Next, the edge prediction error~(EE) measures the quality of classification on the part relations, borrowing the same metric from StructureNet~\cite{10.1145/3355089.3356527}. 
The lower EE is likely to produce a more consistent structure, which means the predicted part boxes are co-related based on their relationships, e.g., symmetry and adjacency.

\begin{table*}[]
\centering
\caption{\textbf{Quantitative Comparison on Structure Inference.}  Please note that AP means part prediction accuracy (\%) computed by average precision with IoU threshold 0.25, and EE means edge prediction error. 
The bold text is used for the best results for each column. 
The columns for key components describe which prior knowledge each method takes, i.e. segmentation prior or skip connection.
}

\begin{tabular}{c|l|cc|cccccc|cc}
\hline
\multirow{2}{*}{Id} & \multicolumn{1}{c|}{\multirow{2}{*}{Method}} & \multicolumn{2}{c|}{Key Comp.}                        & \multicolumn{2}{c}{Chair} & \multicolumn{2}{c}{Table} & \multicolumn{2}{c|}{Storage}     & \multicolumn{2}{c}{Average} \\ \cline{3-12} 
                    & \multicolumn{1}{c|}{}                        & Seg.                      & Skip.                     & AP (\%)         & EE (↓)  & AP (\%)      & EE (↓)     & AP (\%)        & EE (↓)         & AP (\%)         & EE (↓) \\ \hline
1                   & $\mathcal{F}_s + \mathcal{G}_{SN}$           &                           &                           & 5.03            & 0.682   & 2.02         & 0.827      & 1.07           & 0.649          & 2.71            & 0.720  \\
2                   & $\mathcal{F} + \mathcal{G}_{SN}$             & \checkmark &                           & 10.79           & 0.421   & 1.28         & 0.786      & 1.95           & \textbf{0.519} & 4.68            & 0.576  \\
3                   & $\mathcal{F} + \mathcal{G}$ (Ours)           & \checkmark & \checkmark & 48.41           & \textbf{0.273}   & 26.36        & 0.440      & 21.57          & 0.693          & 32.11           & \textbf{0.469}  \\
4                   & Ours $+ \mathcal{M}$                         & \checkmark & \checkmark & \textbf{48.86}  & \textbf{0.273}   & \textbf{26.82}        & \textbf{0.436}      & \textbf{22.08} & 0.697          & \textbf{32.59}  & \textbf{0.469}  \\ \hline
\end{tabular}
\label{tab:results_forward}
\vspace{-2mm}
\end{table*}

\textbf{Results.}
We conducted the qualitative and quantitative evaluations and the results are summarized in Figure~\ref{fig:results_forward} and Table~\ref{tab:results_forward}.
We first illustrate the results of the qualitative evaluation in Figure~\ref{fig:results_forward}.
As illustrated, our method predicts the most plausible structures for all categories even for the cases of complex input geometry, resembling the target structures compared to other baselines.
However, other baselines almost deliver inaccurate appearance even from given input shape and less realistic structure.
Even in the case of an approximate structure predicted, they fail to recover the precise part geometry of small and thin parts.
The second baseline, in particular, also fails to transfer the information of part regions extracted to the decoding step, although clear correspondence exists.
Here, we can observe that naive encoder-decoder methods cannot deliver not only accurate structure prediction but also the diversity of output, where the predicted part structure has a similar appearance from apparently different input shapes.
Therefore, we argue that fully utilizing part segmentation priors and the point-to-part association clearly helps the structure inference.

Next, we demonstrate our quantitative evaluation results in Table~\ref{tab:results_forward}. 
Obviously, ours outperforms the naive encoder-decoder baselines in two metrics, leaving significant margins for both.
The numbers in the first row describe the results of the first baseline without having any priors that our method uses, i.e., part segmentation and point-to-part association, which almost fails to predict the accurate structures.
The other baseline also does not achieve good results either, only with a small improvement from the first baseline.
Although this one takes the hierarchically aggregated latent code using both segmentation and hierarchy priors, the structure prediction accuracy slightly increased. 
By comparing it with ours, we find utilizing the priors in our method helpful showing a significant margin of 27.43\% in part prediction accuracy and 0.107 in edge prediction error.
Moreover, ours leaves more margin by 27.91\% for the refined structure after the segmentation refinement (bottom row).
We will discuss this later~(Sec~\ref{exp_backward}).


\subsection{Evaluation on Segmentation Refinement}\label{exp_backward}
\textbf{Baselines and Metric.}
We compared our method to the other state-of-the-art part segmentation methods, covering SGPN~\cite{wang2018sgpn}, PartNet~\cite{mo2019partnet}, Probabilistic Embedding~(PE)~\cite{zhang2021point}, and PointGroup~\cite{jiang2020pointgroup}.
To show the necessity of our proposed method using structural context, we compare ours to another simple baseline that predicts merge operation directly from the output from part segmentation without using any structural priors.
The candidates are detected using the part bounding boxes from a principal component analysis~(PCA)-based oriented bounding box estimator. 
As same as the introduced methods, the quantitative evaluation is performed based on a class-wise \emph{mean Average Precision (mAP)} with IoU threshold 0.5. 

\textbf{Results.}
We discuss how the proposed method refines the given part segmentation output with quantitative and qualitative evaluations.
Please note that we will focus on the \emph{improvement} on the segmentation quality from our backbone $\psi$ since we do not train any additional part segmentation network in our framework.

\begin{figure}[]
    \centering
    \includegraphics[width=0.48\textwidth]{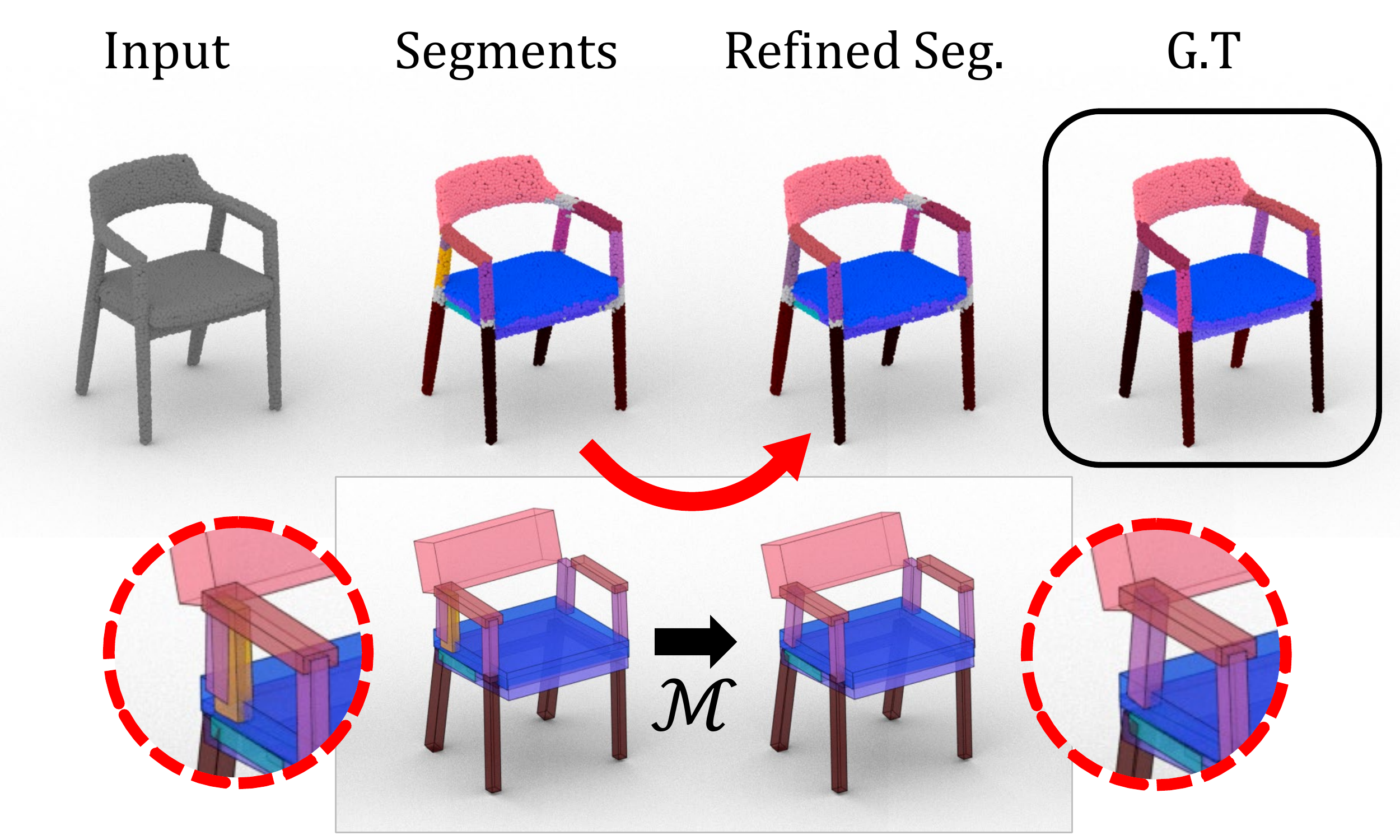}
    \caption{\textbf{Structure-to-Segmentation Refinement.} 
    Utilizing structural information, ours rectify the first segmentation through refinement network $\mathcal{M}$. 
    The circle describes a closer look at the region of conflict in the predicted part structure. 
    After merge prediction, we can get refined segmentation updating noisy regions.
    }
    \label{fig:refine}
    \vspace{-3mm}
\end{figure}

\begin{figure} 
    \centering
    \includegraphics[width=0.47\textwidth]{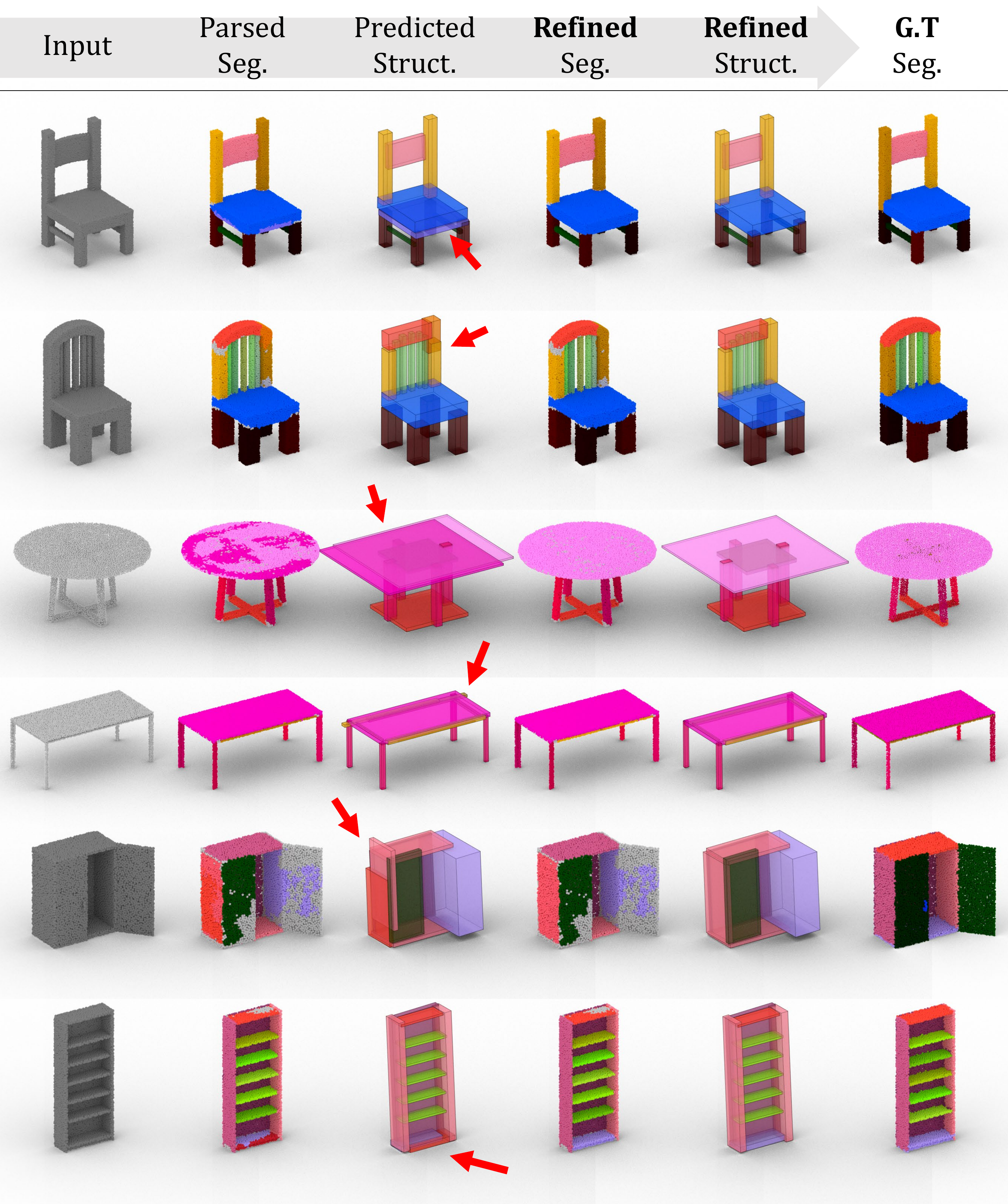}
    \caption{\textbf{The Interplay between Part Segmentation and Structure Inference.} The ground-truth part segmentation is at the right-most column. 
    We point the region of conflict using a red arrow. 
    }
    \label{fig:refine_all}
\end{figure}

In Figure~\ref{fig:refine}, we illustrate the visualization of this refinement process for clearance to demonstrate how the merge operation gives us the refined part segmentation.
Given an initial segmentation output and the prediction structure from it, we first predict merge operation by $\mathcal{M}$ for the candidate nodes detected by the conflict of the part boxes~(indicated by the red arrow).
By incorporating the set of structure-aware features, ours refines the first part segmentation result by merging the candidate part segment to its target segment.  
From the visuals, we observe the accurate merge prediction gives us more realistic and clear part segmentation. 
We provide more examples in Figure~\ref{fig:refine_all}.

We demonstrate the result of quantitative evaluation in Table~\ref{tab:refine}.
Before our merge prediction, we observe that PointGroup itself achieves state-of-the-art performance on part instance segmentation overall.
We observe that this quality can be much enhanced after structure-driven merge prediction (bottom row), showing the improved segmentation accuracy by 0.5\% in average.
However, for our compared baseline, we observe that the accuracy rather decreases on average, where the merge prediction is not aware of the structural information. 
For the chair category, the number seems not improved that much since most of the chair part segments are relatively small to make a bigger improvement even though with the correct merge prediction.
On the other hand, the other categories have the bigger improvement where most of the merge cases occur in the bigger part regions, as shown in Figure~\ref{fig:refine_all}. 
Based on the evaluations, we claim that the synergy between the part segmentation and structure inference enables us to improve both tasks by exploiting the supervision for each task, i.e., segmentation and structural priors.

Moreover, we observe this interplay between two tasks further improves the quality of part structure again.
In Table~\ref{tab:results_forward}, we observe that rectified segmentation further can be used to enhance the quality of part structure once again, improved by 0.48\% (bottom row). 
In Figure~\ref{fig:refine_all} (fifth column), we also draw the effectiveness of this refinement enabling our method to achieve more clean and plausible outputs.


\begin{table}[]
\caption{\textbf{Quantitative Results of Segmentation Refinement}. The numbers are calculated by mean average precision (mAP) for each shape category. Compared to the PCA-based baseline which rather decreases the performance, ours with structure-aware features has shown the clear advantage of the proposed method. }
\begin{tabular}{lcccc}
\hline
                                                    & \multicolumn{1}{l}{Avg} & \multicolumn{1}{l}{Chair} & \multicolumn{1}{l}{Stora.} & \multicolumn{1}{l}{Table} \\ \hline
SGPN~\cite{wang2018sgpn}                            & 18.5                    & 19.4                      & 21.5                       & 14.6                      \\
PartNet~\cite{mo2019partnet}                        & 26.8                    & 29.0                      & 27.5                       & 23.9                      \\
PE~\cite{zhang2021point}                            & 31.5                    & 34.7                      & \textbf{34.2}              & 25.5                      \\
PointGroup ($\psi$)~\cite{jiang2020pointgroup}      & 32.0                    & 40.7                      & 26.8                       & 28.5                      \\ \hline
$\psi$ + PCA-box                                    & 31.6                    & 40.7                      & 26.9                       & 27.2                      \\
$\psi$ + Ours                                       & \textbf{32.5}           & \textbf{40.8}             & 27.5                       & \textbf{29.3}             \\ \hline
\label{tab:refine}
\end{tabular}
\vspace{-4mm}
\end{table}

\subsection{Structure-Aware Shape Retrieval}
As an application based on our proposed framework, we introduce a \emph{structure-aware} shape retrieval.
Shape retrieval, which is to search for the most resemble shape in the database given a query shape, has been one of the most practical applications to measure the shape difference.
Currently, there has been a typical and dominant approach to comparing two shapes by measuring a \emph{fitting distance}, which is usually computed by chamfer-distance (CD). 
This yields perceptional failure cases where we seek to find a similar shape in perspective of the semantics and structure.

\begin{figure}
    \centering
    \includegraphics[width=0.47\textwidth]{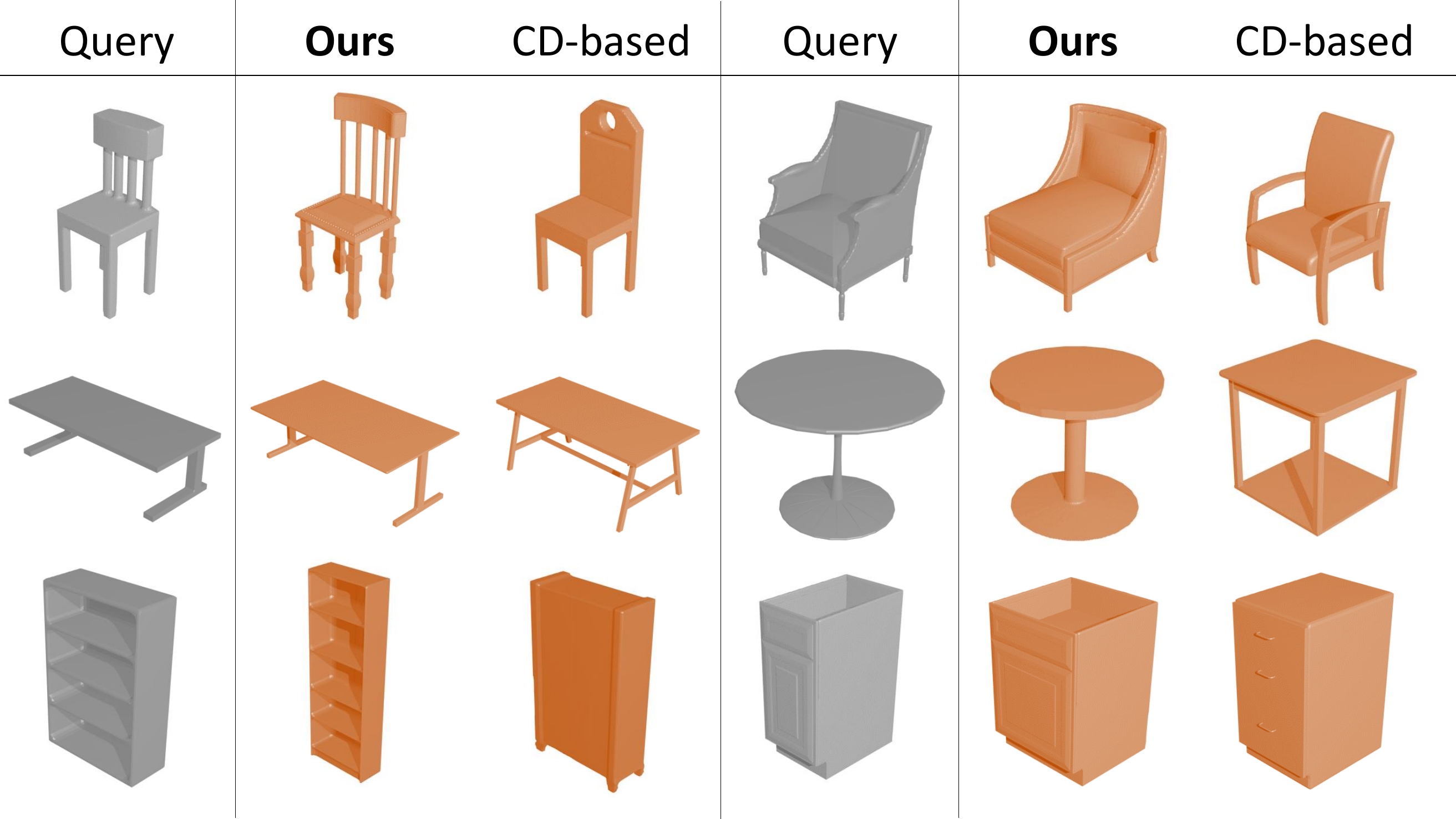}
    \caption{\textbf{The Top-1 Retrieval results of Structure-aware vs. CD-based approaches.}}
    \label{fig:retrieval}
\end{figure}

To tackle this, we propose a structure-driven approach, measuring a \emph{structure difference} that reflects the similarity of semantics between the query shape and shapes in the shape collection.
We showcase the results of top-1 shape retrieval, comparing our structure-aware retrieval with the CD-based method in Figure~\ref{fig:retrieval}. 
Here, we find that our method does not yield the smallest fitting distance, while the retrieved shapes have more similar \emph{semantic} parts.
For more results, we refer readers to the supplementary.

\section{Conclusion}
We proposed \ShortName{}, a framework leveraging the interplay between part segmentation and structure inference in a 3D shape to fully exploit two different supervisions, such as point-to-part associations and hierarchical part structure, and thus improve performance in both tasks making a loop between them. 
Our experimental results demonstrate that this interplay between segmentation and structure inference enables overcoming the performance barrier of existing methods solving only one of the tasks. 
\\

\textbf{Acknowledgements}
This work was supported in part by the NRF grant (No. 2021R1A2C2011459) and NST grant (No. CRC 21011) funded by the Korea government(MSIT). Minhyuk Sung acknowledges the support of grants from Adobe, KT, Samsung Electronics, and ETRI.

{\small
\bibliographystyle{ieee_fullname}
\bibliography{egbib}

\begin{thebibliography}{10}\itemsep=-1pt

\bibitem{attene2006mesh}
Marco Attene, Sagi Katz, Michela Mortara, Giuseppe Patan{\'e}, Michela
  Spagnuolo, and Ayellet Tal.
\newblock Mesh segmentation-a comparative study.
\newblock In {\em IEEE International Conference on Shape Modeling and
  Applications 2006 (SMI'06)}, pages 7--7. IEEE, 2006.

\bibitem{avetisyan2019scan2cad}
Armen Avetisyan, Manuel Dahnert, Angela Dai, Manolis Savva, Angel~X Chang, and
  Matthias Nie{\ss}ner.
\newblock {Scan2CAD}: Learning {CAD} model alignment in {RGB-D} scans.
\newblock In {\em CVPR}, 2019.

\bibitem{10.1145/2185520.2185574}
Martin Bokeloh, Michael Wand, Hans-Peter Seidel, and Vladlen Koltun.
\newblock An algebraic model for parameterized shape editing.
\newblock {\em ACM TOG}, 2012.

\bibitem{chang2015shapenet}
Angel~X Chang, Thomas Funkhouser, Leonidas Guibas, Pat Hanrahan, Qixing Huang,
  Zimo Li, Silvio Savarese, Manolis Savva, Shuran Song, Hao Su, et~al.
\newblock {Shapenet}: An information-rich {3D} model repository.
\newblock {\em CoRR}, abs/1512.03012, 2015.

\bibitem{Chaudhuri2011ProbabilisticRF}
Siddhartha Chaudhuri, Evangelos Kalogerakis, Leonidas~J. Guibas, and Vladlen
  Koltun.
\newblock Probabilistic reasoning for assembly-based {3D} modeling.
\newblock {\em ACM Transactions on Graphics (TOG)}, 30:35, 2011.

\bibitem{chen2009benchmark}
Xiaobai Chen, Aleksey Golovinskiy, and Thomas Funkhouser.
\newblock A benchmark for 3d mesh segmentation.
\newblock {\em Acm transactions on graphics (tog)}, 28(3):1--12, 2009.

\bibitem{chen2020bspnet}
Zhiqin Chen, Andrea Tagliasacchi, and Hao Zhang.
\newblock {BSP-Net}: Generating compact meshes via binary space partitioning.
\newblock In {\em CVPR}, 2020.

\bibitem{dahnert2019joint}
Manuel Dahnert, Angela Dai, Leonidas~J Guibas, and Matthias Nie{\ss}ner.
\newblock Joint embedding of {3D} scan and {CAD} objects.
\newblock In {\em ICCV}, 2019.

\bibitem{ganapathi2018parsing}
Vignesh Ganapathi-Subramanian, Olga Diamanti, Soeren Pirk, Chengcheng Tang,
  Matthias Niessner, and Leonidas Guibas.
\newblock Parsing geometry using structure-aware shape templates.
\newblock In {\em 2018 International Conference on 3D Vision (3DV)}, pages
  672--681. IEEE, 2018.

\bibitem{10.1145/3355089.3356488}
Lin Gao, Jie Yang, Tong Wu, Yu-Jie Yuan, Hongbo Fu, Yu-Kun Lai, and Hao Zhang.
\newblock {SDM-NET}: Deep generative network for structured deformable mesh.
\newblock {\em ACM TOG}, 2019.

\bibitem{10.1145/1409060.1409098}
Aleksey Golovinskiy and Thomas Funkhouser.
\newblock Randomized cuts for {3D} mesh analysis.
\newblock {\em ACM TOG}, 2008.

\bibitem{han2020occuseg}
Lei Han, Tian Zheng, Lan Xu, and Lu Fang.
\newblock Occuseg: Occupancy-aware 3d instance segmentation.
\newblock In {\em Proceedings of the IEEE/CVF conference on computer vision and
  pattern recognition}, pages 2940--2949, 2020.

\bibitem{huang2021multibodysync}
Jiahui Huang, He Wang, Tolga Birdal, Minhyuk Sung, Federica Arrigoni, Shi-Min
  Hu, and Leonidas~J Guibas.
\newblock {MultiBodySync}: Multi-body segmentation and motion estimation via
  {3D} scan synchronization.
\newblock In {\em CVPR}, 2021.

\bibitem{izadinia2017im2cad}
Hamid Izadinia, Qi Shan, and Steven~M Seitz.
\newblock Im2cad.
\newblock In {\em CVPR}, 2017.

\bibitem{jiang2020pointgroup}
Li Jiang, Hengshuang Zhao, Shaoshuai Shi, Shu Liu, Chi-Wing Fu, and Jiaya Jia.
\newblock {PointGroup}: Dual-set point grouping for {3D} instance segmentation.
\newblock In {\em CVPR}, 2020.

\bibitem{jones2020shapeassembly}
R~Kenny Jones, Theresa Barton, Xianghao Xu, Kai Wang, Ellen Jiang, Paul
  Guerrero, Niloy~J Mitra, and Daniel Ritchie.
\newblock Shapeassembly: Learning to generate programs for 3d shape structure
  synthesis.
\newblock {\em ACM Transactions on Graphics (TOG)}, 39(6):1--20, 2020.

\bibitem{jonesneurally}
R~Kenny Jones, Aalia Habib, Rana Hanocka, and Daniel Ritchie.
\newblock The neurally-guided shape parser: Grammar-based labeling of 3d shape
  regions with approximate inference.
\newblock In {\em Proceedings of the IEEE/CVF Conference on Computer Vision and
  Pattern Recognition}, 2022.

\bibitem{10.1145/2611811}
Oliver~Van Kaick, Noa Fish, Yanir Kleiman, Shmuel Asafi, and Daniel Cohen-OR.
\newblock Shape segmentation by approximate convexity analysis.
\newblock {\em ACM TOG}, 2015.

\bibitem{Kalogerakis:2017:ShapePFCN}
Evangelos Kalogerakis, Melinos Averkiou, Subhransu Maji, and Siddhartha
  Chaudhuri.
\newblock 3{D} shape segmentation with projective convolutional networks.
\newblock In {\em CVPR}, 2017.

\bibitem{kalogerakis2012probabilistic}
Evangelos Kalogerakis, Siddhartha Chaudhuri, Daphne Koller, and Vladlen Koltun.
\newblock A probabilistic model for component-based shape synthesis.
\newblock {\em ACM Transactions on Graphics (TOG)}, 31(4):1--11, 2012.

\bibitem{kalogerakis2010learning}
Evangelos Kalogerakis, Aaron Hertzmann, and Karan Singh.
\newblock Learning {3D} mesh segmentation and labeling.
\newblock {\em ACM Transactions on Graphics (TOG)}, 29(4):102, 2010.

\bibitem{Kim13}
Vladimir~G. Kim, Wilmot Li, Niloy~J. Mitra, Siddhartha Chaudhuri, Stephen
  DiVerdi, and Thomas Funkhouser.
\newblock Learning part-based templates from large collections of 3d shapes.
\newblock {\em Transactions on Graphics (Proc. of SIGGRAPH)}, 32(4), 2013.

\bibitem{10.1145/2366145.2366157}
Young~Min Kim, Niloy~J. Mitra, Dong-Ming Yan, and Leonidas Guibas.
\newblock Acquiring {3D} indoor environments with variability and repetition.
\newblock {\em ACM TOG}, 2012.

\bibitem{kingma2014adam}
Diederik~P Kingma and Jimmy Ba.
\newblock Adam: A method for stochastic optimization.
\newblock {\em arXiv preprint arXiv:1412.6980}, 2014.

\bibitem{kuhn1955hungarian}
Harold~W Kuhn.
\newblock The hungarian method for the assignment problem.
\newblock {\em Naval research logistics quarterly}, 1955.

\bibitem{li2017grass}
Jun Li, Kai Xu, Siddhartha Chaudhuri, Ersin Yumer, Hao Zhang, and Leonidas
  Guibas.
\newblock Grass: Generative recursive autoencoders for shape structures.
\newblock {\em ACM TOG}, 2017.

\bibitem{lin2017focal}
Tsung-Yi Lin, Priya Goyal, Ross Girshick, Kaiming He, and Piotr Doll{\'a}r.
\newblock Focal loss for dense object detection.
\newblock In {\em ICCV}, 2017.

\bibitem{liu2009part}
Rong Liu, Hao Zhang, Ariel Shamir, and Daniel Cohen-Or.
\newblock A part-aware surface metric for shape analysis.
\newblock In {\em Computer Graphics Forum}, volume~28, pages 397--406. Wiley
  Online Library, 2009.

\bibitem{luo2020learning}
Tiange Luo, Kaichun Mo, Zhiao Huang, Jiarui Xu, Siyu Hu, Liwei Wang, and Hao
  Su.
\newblock {Learning to Group}: A bottom-up framework for {3D} part discovery in
  unseen categories.
\newblock In {\em ICLR}, 2020.

\bibitem{mitra2006partial}
Niloy~J Mitra, Leonidas~J Guibas, and Mark Pauly.
\newblock Partial and approximate symmetry detection for 3d geometry.
\newblock {\em ACM Transactions on Graphics (TOG)}, 25(3):560--568, 2006.

\bibitem{10.1145/3355089.3356527}
Kaichun Mo, Paul Guerrero, Li Yi, Hao Su, Peter Wonka, Niloy~J. Mitra, and
  Leonidas~J. Guibas.
\newblock {StructureNet}: Hierarchical graph networks for {3D} shape
  generation.
\newblock {\em ACM TOG}, 2019.

\bibitem{mo2020structedit}
Kaichun Mo, Paul Guerrero, Li Yi, Hao Su, Peter Wonka, Niloy~J Mitra, and
  Leonidas~J Guibas.
\newblock {StructEdit}: Learning structural shape variations.
\newblock In {\em CVPR}, 2020.

\bibitem{mo2021where2act}
Kaichun Mo, Leonidas~J Guibas, Mustafa Mukadam, Abhinav Gupta, and Shubham
  Tulsiani.
\newblock {Where2Act}: From pixels to actions for articulated {3D} objects.
\newblock In {\em ICCV}, 2021.

\bibitem{mo2020pt2pc}
Kaichun Mo, He Wang, Xinchen Yan, and Leonidas Guibas.
\newblock {PT2PC}: Learning to generate {3D} point cloud shapes from part tree
  conditions.
\newblock In {\em ECCV}, 2020.

\bibitem{mo2019partnet}
Kaichun Mo, Shilin Zhu, Angel~X Chang, Li Yi, Subarna Tripathi, Leonidas~J
  Guibas, and Hao Su.
\newblock {Partnet}: A large-scale benchmark for fine-grained and hierarchical
  part-level {3D} object understanding.
\newblock In {\em CVPR}, 2019.

\bibitem{muller2006procedural}
Pascal M{\"u}ller, Peter Wonka, Simon Haegler, Andreas Ulmer, and Luc Van~Gool.
\newblock Procedural modeling of buildings.
\newblock In {\em ACM SIGGRAPH 2006 Papers}, pages 614--623. 2006.

\bibitem{10.1145/2366145.2366156}
Liangliang Nan, Ke Xie, and Andrei Sharf.
\newblock A search-classify approach for cluttered indoor scene understanding.
\newblock {\em ACM TOG}, 2012.

\bibitem{niu2018im2struct}
Chengjie Niu, Jun Li, and Kai Xu.
\newblock {Im2struct}: Recovering {3D} shape structure from a single {RGB}
  image.
\newblock In {\em CVPR}, 2018.

\bibitem{10.1145/2010324.1964928}
Maks Ovsjanikov, Wilmot Li, Leonidas Guibas, and Niloy~J. Mitra.
\newblock Exploration of continuous variability in collections of {3D} shapes.
\newblock {\em ACM TOG}, 2011.

\bibitem{Paschalidou2020CVPR}
Despoina Paschalidou, Luc van Gool, and Andreas Geiger.
\newblock Learning unsupervised hierarchical part decomposition of 3d objects
  from a single rgb image.
\newblock In {\em Proceedings IEEE Conf. on Computer Vision and Pattern
  Recognition (CVPR)}, June 2020.

\bibitem{paszke2017automatic}
Adam Paszke, Sam Gross, Soumith Chintala, Gregory Chanan, Edward Yang, Zachary
  DeVito, Zeming Lin, Alban Desmaison, Luca Antiga, and Adam Lerer.
\newblock Automatic differentiation in pytorch.
\newblock 2017.

\bibitem{qi2016pointnet}
Charles~R Qi, Hao Su, Kaichun Mo, and Leonidas~J Guibas.
\newblock {PointNet}: Deep learning on point sets for {3D} classification and
  segmentation.
\newblock In {\em CVPR}, 2017.

\bibitem{qi2017pointnetplusplus}
Charles~R Qi, Li Yi, Hao Su, and Leonidas~J Guibas.
\newblock {PointNet++}: Deep hierarchical feature learning on point sets in a
  metric space.
\newblock In {\em CVPR}, 2017.

\bibitem{shamir2008survey}
Ariel Shamir.
\newblock A survey on mesh segmentation techniques.
\newblock In {\em Computer graphics forum}, volume~27, pages 1539--1556. Wiley
  Online Library, 2008.

\bibitem{song2016deep}
Shuran Song and Jianxiong Xiao.
\newblock Deep sliding shapes for amodal {3D} object detection in {RGB-D}
  images.
\newblock In {\em CVPR}, 2016.

\bibitem{10.1145/3414685.3417783}
Minhyuk Sung, Zhenyu Jiang, Panos Achlioptas, Niloy~J. Mitra, and Leonidas~J.
  Guibas.
\newblock {DeformSyncNet}: Deformation transfer via synchronized shape
  deformation spaces.
\newblock {\em ACM TOG}, 2020.

\bibitem{sung2015data}
Minhyuk Sung, Vladimir~G Kim, Roland Angst, and Leonidas Guibas.
\newblock Data-driven structural priors for shape completion.
\newblock {\em ACM TOG}, 2015.

\bibitem{10.1145/3130800.3130821}
Minhyuk Sung, Hao Su, Vladimir~G. Kim, Siddhartha Chaudhuri, and Leonidas
  Guibas.
\newblock {ComplementMe}: Weakly-supervised component suggestions for {3D}
  modeling.
\newblock {\em ACM TOG}, 2017.

\bibitem{Tian2019LearningTI}
Yonglong Tian, Andrew Luo, Xingyuan Sun, Kevin Ellis, William~T. Freeman,
  Joshua~B. Tenenbaum, and Jiajun Wu.
\newblock Learning to infer and execute {3D} shape programs.
\newblock 2019.

\bibitem{uy2021joint}
Mikaela~Angelina Uy, Vladimir~G Kim, Minhyuk Sung, Noam Aigerman, Siddhartha
  Chaudhuri, and Leonidas~J Guibas.
\newblock Joint learning of {3D} shape retrieval and deformation.
\newblock In {\em CVPR}, 2021.

\bibitem{10.1145/2461912.2461924}
Oliver van Kaick, Kai Xu, Hao Zhang, Yanzhen Wang, Shuyang Sun, Ariel Shamir,
  and Daniel Cohen-Or.
\newblock Co-hierarchical analysis of shape structures.
\newblock {\em ACM TOG}, 2013.

\bibitem{vu2022softgroup}
Thang Vu, Kookhoi Kim, Tung~M. Luu, Xuan~Thanh Nguyen, and Chang~D. Yoo.
\newblock {SoftGroup} for {3D} instance segmentation on {3D} point clouds.
\newblock In {\em CVPR}, 2022.

\bibitem{wang2018sgpn}
Weiyue Wang, Ronald Yu, Qiangui Huang, and Ulrich Neumann.
\newblock {SGPN}: Similarity group proposal network for {3D} point cloud
  instance segmentation.
\newblock In {\em CVPR}, 2018.

\bibitem{wang2019associatively}
Xinlong Wang, Shu Liu, Xiaoyong Shen, Chunhua Shen, and Jiaya Jia.
\newblock Associatively segmenting instances and semantics in point clouds.
\newblock In {\em Proceedings of the IEEE/CVF Conference on Computer Vision and
  Pattern Recognition}, pages 4096--4105, 2019.

\bibitem{wang2018learning}
Xiaogang Wang, Bin Zhou, Haiyue Fang, Xiaowu Chen, Qinping Zhao, and Kai Xu.
\newblock Learning to group and label fine-grained shape components.
\newblock {\em ACM Transactions on Graphics (SIGGRAPH Asia 2018)}, 37(6), 2018.

\bibitem{wang2012active}
Yunhai Wang, Shmulik Asafi, Oliver Van~Kaick, Hao Zhang, Daniel Cohen-Or, and
  Baoquan Chen.
\newblock Active co-analysis of a set of shapes.
\newblock {\em ACM Transactions on Graphics (TOG)}, 31(6):165, 2012.

\bibitem{wang2013projective}
Yunhai Wang, Minglun Gong, Tianhua Wang, Daniel Cohen-Or, Hao Zhang, and
  Baoquan Chen.
\newblock Projective analysis for 3d shape segmentation.
\newblock {\em ACM Transactions on Graphics (TOG)}, 32(6):1--12, 2013.

\bibitem{wu2019sagnet}
Zhijie Wu, Xiang Wang, Di Lin, Dani Lischinski, Daniel Cohen-Or, and Hui Huang.
\newblock {SAGNet}: Structure-aware generative network for {3D}-shape modeling.
\newblock {\em ACM Transactions on Graphics (TOG)}, 38(4):1--14, 2019.

\bibitem{10.1145/1576246.1531341}
Weiwei Xu, Jun Wang, KangKang Yin, Kun Zhou, Michiel van~de Panne, Falai Chen,
  and Baining Guo.
\newblock Joint-aware manipulation of deformable models.
\newblock In {\em ACM SIGGRAPH}, 2009.

\bibitem{yang2020dsg}
Jie Yang, Kaichun Mo, Yu-Kun Lai, Leonidas~J Guibas, and Lin Gao.
\newblock {DSG-Net}: Learning disentangled structure and geometry for {3D}
  shape generation.
\newblock {\em CoRR}, abs/2008.05440, 2020.

\bibitem{Yi17}
Li Yi, Leonidas Guibas, Aaron Hertzmann, Vladimir~G. Kim, Hao Su, and Ersin
  Yumer.
\newblock Learning hierarchical shape segmentation and labeling from online
  repositories.
\newblock In {\em ACM SIGGRAPH}, 2017.

\bibitem{yi2016scalable}
Li Yi, Vladimir~G Kim, Duygu Ceylan, I-Chao Shen, Mengyan Yan, Hao Su, Cewu Lu,
  Qixing Huang, Alla Sheffer, and Leonidas Guibas.
\newblock A scalable active framework for region annotation in 3d shape
  collections.
\newblock {\em ACM Transactions on Graphics (ToG)}, 35(6):1--12, 2016.

\bibitem{yi2017syncspeccnn}
Li Yi, Hao Su, Xingwen Guo, and Leonidas~J Guibas.
\newblock Syncspeccnn: Synchronized spectral cnn for 3d shape segmentation.
\newblock In {\em Proceedings of the IEEE Conference on Computer Vision and
  Pattern Recognition}, pages 2282--2290, 2017.

\bibitem{yu2019partnet}
Fenggen Yu, Kun Liu, Yan Zhang, Chenyang Zhu, and Kai Xu.
\newblock {Partnet}: A recursive part decomposition network for fine-grained
  and hierarchical shape segmentation.
\newblock In {\em CVPR}, 2019.

\bibitem{zhang2021point}
Biao Zhang and Peter Wonka.
\newblock Point cloud instance segmentation using probabilistic embeddings.
\newblock In {\em CVPR}, 2021.

\bibitem{zhu2018scores}
Chenyang Zhu, Kai Xu, Siddhartha Chaudhuri, Renjiao Yi, and Hao Zhang.
\newblock {SCORES}: Shape composition with recursive substructure priors.
\newblock {\em ACM TOG}, 2018.

\end{thebibliography}
}

\clearpage
\appendix
\section{Appendix}
In the supplementary material, we first introduce more details about the architecture design in our framework (Section~\ref{ArchDesign}). Next, we demonstrate the implementation details and training setup for our framework (Section~\ref{ImpDetails}). Finally, we provide more experimental results to evaluate the proposed architecture including ablation studies~(Section~\ref{MoreResults}). 

\subsection{Architecture Design}\label{ArchDesign}

\subsubsection{Representation of Structure Hierarchy}
The structure used in our framework has two kinds of relationships: \textbf{H} and \textbf{R}, where \textbf{H} $\subset P^2$ describes the directed connectivity between a parent node and its sibling nodes and $\textbf{R}$ is a $M \times M$ matrix that describes the entire part relations between the nodes under the same parent node in the structure.
In the following script, we use $(m_i,m_j,\tau)$ to describe an edge between $m_i$ and $m_j$ with a set of types of part relation $\tau\in\mathcal{T}$. 
The type of part relations $\mathcal{T}$ covers translational, reflective, rotational symmetry, and adjacency, denoted as $\{\tau^{trs},\tau^{ref},\tau^{rot},\tau^{adj}\}$. 
Same as StructureNet~\cite{10.1145/3355089.3356527}, we assume that a part-relation $\tau$ only exists between the nodes in the same subset of the tree and one pair of nodes can have multiple types of relations among $\mathcal{T}$.
The part relation $\tau$ is encoded as a one-hot vector. We do not predict transformation parameters for symmetry relations.

\subsubsection{Parsing-based Structure Encoder} 
\textbf{Structure Tree Construction. }
Given the shape point cloud $A\in\mathbb{R}^{N\times3}$, we first parse the input into a set of part segments $B=\{b_l\}_{l\in L}$, where a part segment $b_l=(X_l,y_l)$ contains a set of points in the corresponding part region $X_l$ and a semantic label $y_l$.
By taking a point cloud of input shape $A$, our backbone $\psi$ decomposes it into a set of leaf part instances. 
Based on the part segmentation output, we get a set of leaf nodes $\{m_l\}_{l \in L}$ for $L$ number of the parts, which will be used for the structure tree construction following.
Here, as mentioned in our paper, we encode each leaf part segment $b_l$ into a 128-dimensional part feature vector~$\textbf{x}_l\in \mathbb{R}^{128}$. 
The feature is encoded by a part feature encoder $f_{part}$, which uses \emph{PointNet++} \cite{qi2017pointnetplusplus} with four set abstraction layers and two linear layers where each layer is followed by ReLU activation and batch normalization. 
\begin{equation}
\label{eq:part feature encoding}
    \textbf{x}_l = f_{part}(X_l) 
\end{equation}

In the tree construction step, we denote a tree node as $\bar{m}_i=(\textbf{x}_i,X_i,y_i)$ for $i \in M$ to distinguish it from the final part node $m_i=(\theta_i, y_i)$. 
Based on the part node extracted in the previous step, we operate node grouping recursively until it reaches to root node $\bar{m}^{root}$.

\begin{figure}[t]
    \centering
    \includegraphics[width=0.48\textwidth]{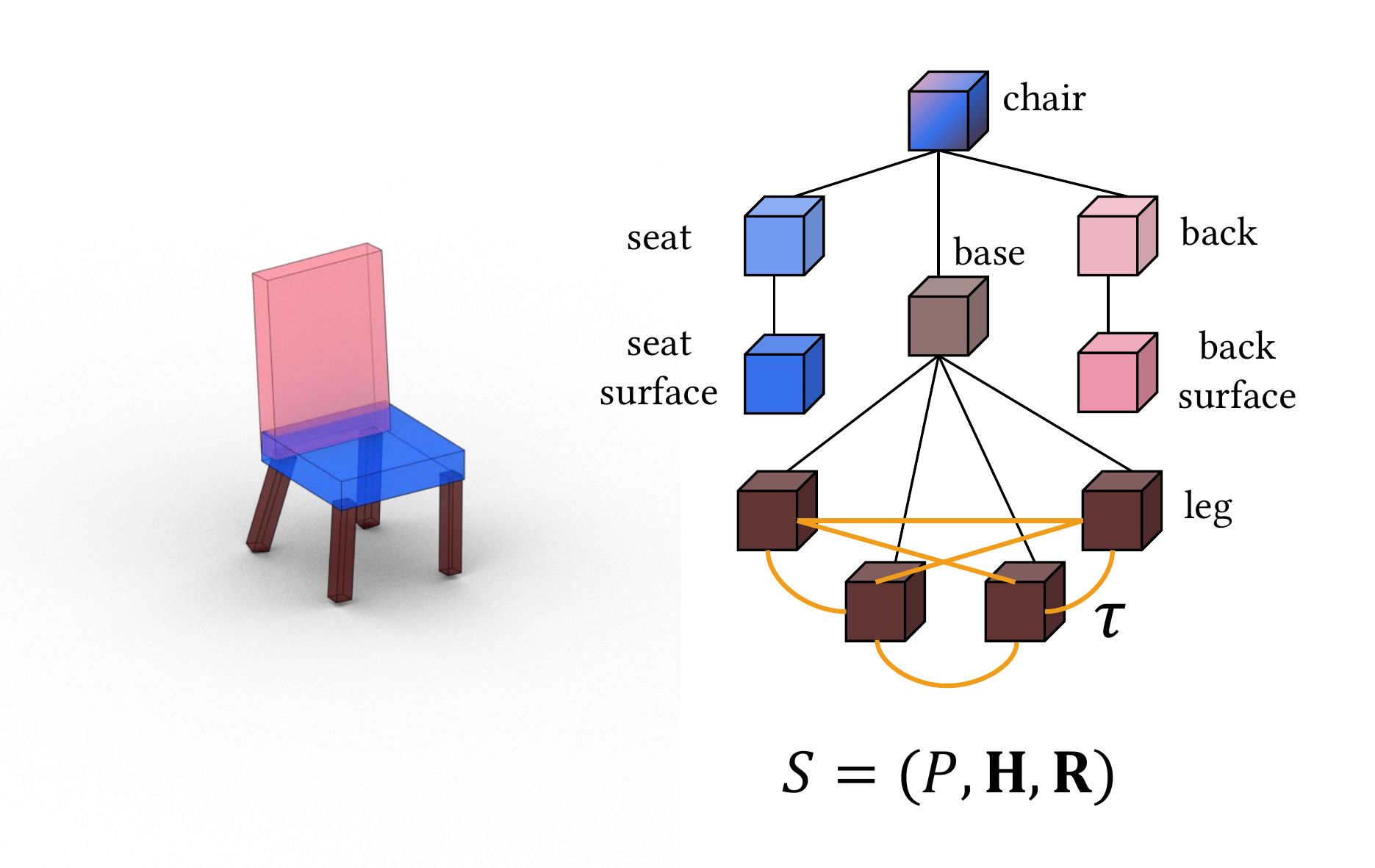}
    \caption{\textbf{Structure Hierarchy Representation.} We represent the part structure~(left) in the tree hierarchy~(right). 
    For each subset in the tree, a parent node and sibling nodes are connected hierarchically in $\textbf{H}$ (black edges) and the sibling nodes have relationships $\tau$ in $\textbf{R}$ (orange edges). The part node in $P$ comprises the oriented bounding box parameters $\theta$ and the semantic label. }
    \label{fig:hierarchy}
\end{figure}

To group a subset of nodes at each level of the hierarchy, we take a heuristic approach based on their semantic labels and spatial distances. 
Since the semantic label $y_i$ describes the level of the tree where the part node belongs to by itself as defined in StructureNet~\cite{10.1145/3355089.3356527}, we construct a structure tree based on them.
For instance, a leaf part with a semantic label named \textit{leg} guarantees that it has to be grouped to a \textit{base} node (Fig.~\ref{fig:hierarchy}). 
We refer the readers to the StructureNet paper for more details about the part annotation for the hierarchy.

In general, the node grouping based on semantic prior works fine. 
For some cases where depending on the semantic information yields an ambiguity, we apply the simple heuristics based on the spatial information of the part points $X_l$.
For example, \textit{arms} in a \textit{chair} can be separated according to the global center coordinates relative to the center of the input shape. 
Here, we can approximate the center of each node by its point cluster $X_l$ by averaging their global coordinates.

\textbf{Hierarchical Feature Encoding. }
Once the hierarchical connectivity $\textbf{H}$ is established, we calculate the encode a parent node's feature vector $\textbf{x}_r$ by aggregating the features of its sibling nodes $\{\textbf{x}_c\}_{c \in C_r}$, where $C_r$ is the number of the sibling nodes.
For the parent feature encoding, we use a feature encoder $f_{child}$ with one single linear layer that takes a set of features from child nodes:: 
\begin{equation}
\label{eq:child feature encoding}
    \textbf{x}_r = f_{child}(\{\textbf{x}_c\})
\end{equation}
In this process, we encourage the parent features to have a broader view of the subset's sibling nodes.
We recursively iterate this process until it approaches to the root feature $\textbf{x}^{root}$.
Finally, we treat this root feature as an \emph{global context} throughout the entire structure, both in the decoding step and merge prediction network later.
As aggregate the node features, we also gather the part point cloud to define a part region for each parent node.

After the construction step, we get all the part nodes and hierarchical connectivity $(\bar{P},\textbf{H})$, where the part nodes $\{\bar{m}_i\}\in\bar{P}$ involve the global context.
We further demonstrate how can we take advantage of these context features and the hierarchy prior to support the structure decoding step in the following.

\subsubsection{Multi-level Context-Aware Structure Decoding} \label{4.2} 
In this section, we introduce the part structure decoder $\mathcal{G}$ using multi-level context across the structure hierarchy in \emph{top-down} manner.
The goal of our part structure decoding is two-fold: regressing the part bounding box $\theta_i$ and classifying the part relationship among the nodes $(m_i,m_j,\tau)$.
Though it is possible to directly decode them directly from segmentation output, we found this yields an inconsistent aligned part structure.
This is mainly due to the nature of 3D shape, the parts in the structure are strongly co-related not only horizontally, but also the vertical~(hierarchical) way.

\begin{figure}
    \centering
    \includegraphics[width=0.49\textwidth]{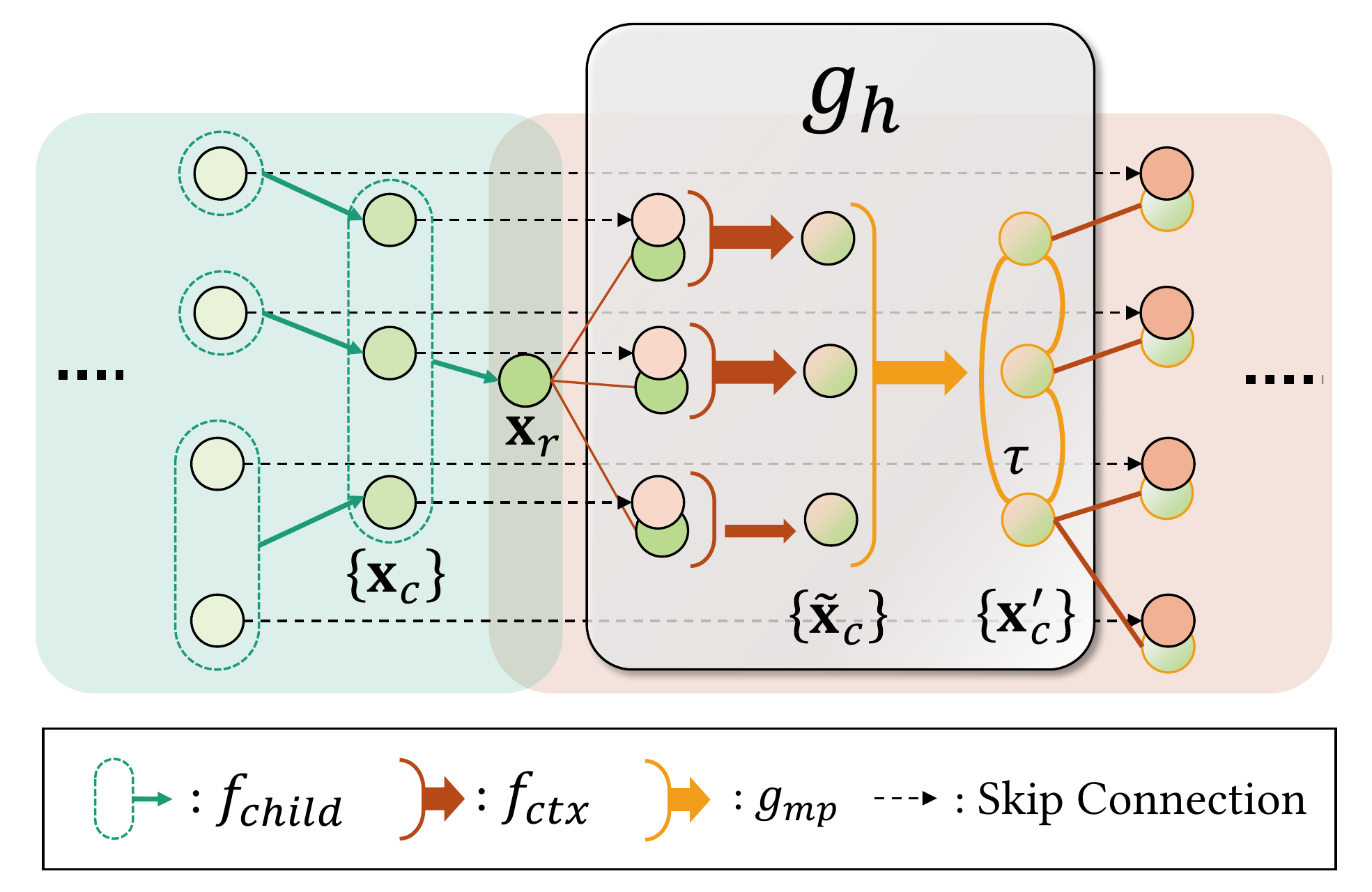}
    \caption{\textbf{The illustration of Hierarchical Message Passing.} Given the constructed hierarchy and all the features encoded using $f_{child}$, we perform hierarchical message passing $g_h$ to make features learn a multi-level context in the structure. In global context learning $f_{ctx}$, the parent feature $\textbf{x}_r$ and the sibling feature connected via a skip connection $\textbf{x}_c$ are aggregated. Then, the local part message passing $g_{mp}$ classifies a part relation $\tau$ and performs inter-part communication between connected sibling nodes. The multi-level context-aware node feature $\textbf{x}'_c$ is again used for the recursive message passing to its sibling subset.}
    \label{fig:hierarchical message passing}
\end{figure}

To tackle this, we propose a hierarchical message passing $g_h$, to learn the structural context based on the \emph{hierarchy prior}. 
Our previous segmentation-based structure construction enables us to map the part regions in the geometry space into the hierarchical representation. 
Based on this, we build an association between part segments and part nodes in the form of a \emph{skip connection} connecting the corresponding parts (Fig~\ref{fig:hierarchical message passing}).  
By fully exploiting both hierarchical connectivity and the association from the skip connection, we perform two types of message passing: 1) global context learning and 2) local part relation learning.

\textbf{Global Context Learning. }                  
The main purpose of this step is to make our part features aware of the global context across the constructed hierarchy in \emph{vertical} way.
Here, the feature vector of each node $\textbf{x}_i$ is updated through parent feature aggregation before learning inter-part relationships.  
This ensures that the part feature not only contains the local relationship between neighboring part nodes but also covers the global context across the different levels of structures. 
Starting from the root node feature $\textbf{x}^{root}$, we propagate each level of global context by decoding it down to the sibling nodes. 

To do this, we use $\textbf{x}_c$ using $f_{ctx}$, a single linear layer that consumes a concatenated feature vector of child and parent nodes $[\textbf{x}_{c};\textbf{x}_{r}] \in \mathbb{R}^{256}$: 
\begin{equation}
\label{eq:global context aggregation}
   \tilde{\textbf{x}}_c = f_{ctx}([\textbf{x}_c;\textbf{x}_r])
\end{equation}
Here, we consider $\tilde{\textbf{x}}_c$ as a global context-aware feature, which supports the part-relation learning and the bounding box decoding further.

\textbf{Local Part Relation Learning. } 
After aggregating the global context to each part node, we perform a \emph{local} message passing to learn an inter-part communication for each subset \emph{horizontally}.

Inspired by StructureNet~\cite{10.1145/3355089.3356527}, we aim to make the part nodes learning predefined local part relations, i.e. symmetry and adjacency by classifying the accurate type of part relation $\tau\in\mathcal{T}$.
For $\mathcal{T}$, we consider three types of symmetry $\{\tau^{trs},\tau^{rot},\tau^{ref}\}$ and the adjacency $\tau^{adj}$ as mentioned earlier. 
With the accurately predicted part relations, the co-related part features can achieves more consistent part geometry prediction further, e.g. preserving a \emph{reflective} symmetry between \textit{arms} in a \textit{chair}.   
Analogous to StructureNet. we connect the whole part nodes inside the subset to each other and compute a set of corresponding edge feature vectors for each pair of nodes through the message passing.
To compute the edge feature $\textbf{y}_{ij}\in\mathbb{R}^{256}$, we use $g_{edge}$, a two-layer multi-layer perceptron that takes two context-aware feature vectors $\{\tilde{\textbf{x}}_i,\tilde{\textbf{x}}_j\}$:
\begin{equation}
\label{eq:edge feature encoding}
    \textbf{y}_{ij} = g_{edge}([\tilde{\textbf{x}}_i;\tilde{\textbf{x}}_j])
\end{equation}

Then we classify which types of relationships exist between them among the part-relations $\mathcal{T}$ for the number of types $|\mathcal{T}|$ given the edge feature $\textbf{y}_{ij}$.
\begin{equation}
\label{eq:edge classification}
    (p_{(\bar{m}_i, \bar{m}_j, \tau_1)},...,p_{(\bar{m}_i, \bar{m}_j, \tau_{|\mathcal{T}|})}) = \sigma(g_{\tau}(\textbf{y}_{ij}))
\end{equation}
where $\sigma$ and $p$ are a sigmoid and a probability function, respectively.
Note that we do not predict any parameters for each relation here. 
Here, we treat edge prediction as a classification problem where we predict the probability of edge types $\in[0,1]$. 

Finally, we perform an inter-part message passing, adopted from StructureNet~\cite{10.1145/3355089.3356527}.
We prune out the edges with a probability lower than the edge threshold, i.e. 0.5, to keep a valid set of edges. 
Next, we perform two iterations of message passing for each part feature with the connected neighbors in the subset, using the edge features with the classified relationships. 
Lastly, we aggregate the features from each iteration to get the finalized node feature $\textbf{x}'_i$ from a set of connected parts $\{\bar{m}_k\}_{k\in K_i}$ where $K_i$ is the number of parts connected.
We refer the readers to the paper of StructureNet~\cite{10.1145/3355089.3356527} for more details about the message-passing network.

Through hierarchical message passing across the subsets and the overall structure, we encourage the final part representation $\textbf{x}'_i$ to incorporate much broader context across the structure beyond a geometry-dependent feature vector.  

\textbf{Bounding Box Decoding. } 
Finally, we regress the bounding box parameters $\theta_i = (\textbf{t}_i,\textbf{s}_i,q_i)$ for each part node in the hierarchy.
Here, we use $g_{box}$ built with a series of linear layers to output the bounding box parameters.
Empirically, we observe that the decoder often fails to perform accurate regression if it works on the feature vector alone.
To address this, we utilize the part points $X_i$ for each node to support the box decoding, which we gathered during the tree construction.
For the translation vector $\textbf{t}_i$, we first compute the rough center coordinates $\hat{\textbf{t}}_i$ by averaging the global coordinates of $X_i$. 
The decoder $g_{box}$ predicts the offset vector $\textbf{o}_i \in \mathbb{R}^3$ to slightly tune the center coordinate as $\textbf{t}_i = \hat{\textbf{t}}_i + \textbf{o}_i$.
By doing so, we can achieve a much precise estimation of box parameters, not only the box center but the others as well.
Similarly, we have seen that it is enough to approximate the scaling vector $\textbf{s}_i$ by examining the volume of part points along the rotation axis prediction done for unit quaternion $q$.
Our final formulation for box parameter decoding is as follows:
\begin{equation}
\label{eq:bounding box decoder}
   (\textbf{t}_i, \textbf{s}_i, q_i) = g_{box}(\textbf{x}'_i,X_i)
\end{equation}
Finally, we get a part node $m_i=(\theta_i,y_i)$ for all the nodes in the structure and complete the part structure inference. 

\subsubsection{Training and Losses. }

\textbf{Structure Inference Network.} 
The parameters for the encoder $\mathcal{F}$ and the decoder $\mathcal{G}$ are supervised both at the structure decoding step through backpropagation. 
As mentioned in our paper, our overall loss design for structure inference network is mostly brought from StructureNet~\cite{10.1145/3355089.3356527} since ours and StructureNet share the equivalent objective, inferring the structure output.
Similar to StructureNet, our total loss $\mathcal{L}_{total}$ is computed by the following steps.
First, a correspondence map $\textbf{M}\subset P \times \hat{P}$ between the predicted parts~$\hat{P}$ and the target parts~$P$ is established to supervise network parameters based on the geometric distance between the parts using the hungarian algorithm~\cite{kuhn1955hungarian}.
Second, the geometry reconstruction error is computed by summing up $\mathcal{L}_{box}$ and $\mathcal{L}_{norm}$, a bounding box fitting distance and an additional normal error for the precise orientation regression, respectively.
Third, we compute an edge prediction loss $\mathcal{L}_{edge}$, where we consider the edge prediction as a binary classification problem, and the classifier is supervised using cross-entropy loss.

Finally, we calculate a structure consistency loss $\mathcal{L}_{cons}$ to make the relationship at parent nodes consistent with their sibling subsets.
For more detail, we refer the readers to the StructureNet paper~\cite{10.1145/3355089.3356527}.
In summary, we compute our total loss as follows:
\begin{equation}
    \mathcal{L}_{total} = \lambda_1\mathcal{L}_{box} +  \lambda_2\mathcal{L}_{norm} + \mathcal{L}_{edge} + \mathcal{L}_{cons} 
\end{equation}
where we define the coefficients $\lambda_1$ and $\lambda_2$ as 20 and 10, respectively.

\textbf{Segmentation Refinement Network.} 
To train part segmentation refinement network $\mathcal{M}$, we use focal loss~\cite{lin2017focal} as discussed in our main paper.
We define the focal loss $F$ taking a probability score $p_s \in [0,1]$ and a target label $p$:
\begin{equation}
\label{eq:focal loss}
    F(p_s, p) = -\alpha_t(1-p_t)^\gamma \log(p_t)
\end{equation}
where $p_t$ is $p_s$ if $p=1$ and $p_t$ is $1-p_s$ otherwise. 
$\alpha_t$ is a weight and $\gamma$ is a focusing parameter.
Empirically, we opt to set $\alpha$ as 0.15 and $\gamma$ as 2.
The merge loss $\mathcal{L}_{merge}$ is then calculated with ground-truth assignment $\bar{\textbf{C}}$ as follows:
\begin{equation}
\label{eq:merge loss}
    \mathcal{L}_{merge} = \sum_{(i,j)\in\bar{\textbf{C}}}F(p^{merge}_{(m_i, m_j)},\mathbbm{1}_{\bar{\textbf{C}}}(i,j))
\end{equation}
where $\mathbbm{1}_{\bar{\textbf{C}}}$ is an indicator function that gives $1$ for the valid candidate pair in $\bar{\textbf{C}}$ and the others $0$.

\begin{table*}[]
\centering
\caption{\textbf{Comparison on Structure Inference.} Same as our paper, AP means part prediction accuracy (\%) computed by average precision with IoU threshold 0.25, and EE means edge prediction error calculated by one minus F1-score of edge prediction outputs. 
Please note that the second and third baselines do not measure EE since they do not predict any part relationships. 
The columns for key components describe which prior knowledge or the level of message passing each method takes.
The bold text means the best results for each column. 
}
\resizebox{\textwidth}{!}{
\begin{tabular}{ll|ccccc|cccccccc}
\hline
\multicolumn{1}{c}{\multirow{3}{*}{Id}} & \multicolumn{1}{c|}{\multirow{3}{*}{Method}}     & \multicolumn{5}{c|}{Key Components}                                                           & \multicolumn{6}{c}{Categories}                                                           & \multicolumn{2}{c}{\multirow{2}{*}{Avg}} \\ \cline{3-13}
\multicolumn{1}{c}{}                    & \multicolumn{1}{c|}{}                            & \multicolumn{2}{c|}{Prior}                       & \multicolumn{3}{c|}{Message Passing}       & \multicolumn{2}{c}{Chair} & \multicolumn{2}{c}{Table} & \multicolumn{2}{c}{Storagefurn.} & \multicolumn{2}{c}{}                     \\ \cline{3-15} 
\multicolumn{1}{c}{}                    & \multicolumn{1}{c|}{}                            & Seg.         & \multicolumn{1}{c|}{Hier.}        & Skip.        & Local        & global       & AP (\%)      & EE (↓)     & AP (\%)      & EE (↓)     & AP (\%)         & EE (↓)         & AP (\%)             & EE (↓)             \\ \hline
1                                       & $\mathcal{F}_s + \mathcal{G}_{SN}$               &              & \multicolumn{1}{c|}{\checkmark} &              & \checkmark & \checkmark & 5.03         & 0.6824     & 2.02         & 0.8272     & 1.07            & 0.6491         & 2.71                & 0.7196             \\ \hline
2                                       & $\psi + \text{PCA}$                              & \checkmark & \multicolumn{1}{c|}{}             &              &              &              & 37.32        & -          & 20.96        & -          & 17.75           & -              & 25.34               & -                  \\
3                                       & $\psi + g_{box}$                                 & \checkmark & \multicolumn{1}{c|}{}             &              &              &              & 46.66        & -          & 25.89        & -          & 19.96           & -              & 30.83               & -                  \\
4                                       & $\psi + g_{box}+g_{mp}$                          & \checkmark & \multicolumn{1}{c|}{}             &              & \checkmark &              & 48.39        & 0.8576     & 24.19        & 0.8835     & 19.96           & 0.8933         & 30.85               & 0.8781             \\ \hline
5                                       & $\mathcal{F} + \mathcal{G}_{SN}$                 & \checkmark & \multicolumn{1}{c|}{\checkmark} &              & \checkmark & \checkmark & 10.79        & 0.4211     & 1.28         & 0.7863     & 1.95            & \textbf{0.5191}         & 4.68                & 0.5755             \\
6                                       & $\mathcal{F} + \mathcal{G} - f_{ctx}$            & \checkmark & \multicolumn{1}{c|}{\checkmark} & \checkmark &              & \checkmark & 47.21        & 0.3006     & 22.91        & 0.4597     & 19.25           & 0.6664         & 29.79               & 0.4756             \\
7                                       & $\mathcal{F} + \mathcal{G} - g_{mp}$             & \checkmark & \multicolumn{1}{c|}{\checkmark} & \checkmark & \checkmark &              & 47.34        & 0.3448     & \textbf{26.41}        & 0.5024     & 21.20            & 0.6867         & 31.65               & 0.5113             \\
8                                       & $\mathcal{F} + \mathcal{G}$ (Ours)                     & \checkmark & \multicolumn{1}{c|}{\checkmark} & \checkmark & \checkmark & \checkmark & \textbf{48.41}        & \textbf{0.2727}     & 26.36        & \textbf{0.4400}       & \textbf{21.57}           & 0.6934         & \textbf{32.11}               & \textbf{0.4687}             \\ \hline
\end{tabular}
}
\label{tab:comparison}
\end{table*}

\subsection{Implementation Details}\label{ImpDetails}
We implement our framework in PyTorch~\cite{paszke2017automatic}. 
The training for each category is performed until convergence with batch size 16, mostly requiring 1-2 days for structure inference network and less than two hours for segmentation refinement network on a single GeForce RTX 3090 and an Intel Xeon Silver 4210R CPU.
We use the Adam optimizer~\cite{kingma2014adam} with the initial learning rate as $0.5^{-3}$ for inference network and $10^{-4}$ for refinement network, decayed by 0.8 per 500 steps. 

\subsection{More Experimental Results}\label{MoreResults}
\subsubsection{Ablation Study on Structure Inference Network}\label{exp:ablation}
To demonstrate the effectiveness of the key components in our proposed hierarchical message passing $g_h$, we perform sets of ablation studies. 
For other methods not explained here, we refer readers to our main paper.

\textbf{Baselines.}
Here, we built another type of baseline, which directly decodes the bounding box from extracted part segments without having hierarchical priors or key components used in our method. 
Based on our backbone $\psi$ and the part feature encoder $f_{part}$, the box decoder baseline parses the input shape into leaf part instances, encodes part features, and directly predicts bounding box parameters~$\{\theta_l\}_{l\in L}$. 
Here, we prepare three kinds of box decoder baselines ($2^{nd}-4^{th}$ rows in Table~\ref{tab:comparison}) directly consuming the output from backbone $\psi$: a PCA-based bounding box estimator, a box decoder $g_{box}$, and the box decoder with message passing network $g_{mp}$.
Since these box encoders do not predict and learn any part relations except the last one, the edge prediction error is not measured.

Note that the last one uses the reduced version of our part-relation learning where the relationships are learned across all the leaf instances, which makes the edge prediction a lot more difficult.
Then, we show how the level of feature updates in the local part-relation learning and global context learning affects the performance of structure inference for each.
To this end, we built baselines by subtracting each component of $g_h$ from our framework~$\mathcal{F}+\mathcal{G}$, represented as the one without local part message passing $g_{mp}$ and global context encoding $f_{ctx}$. 

\begin{figure*}
    \centering
    \includegraphics[width=1.01\textwidth]{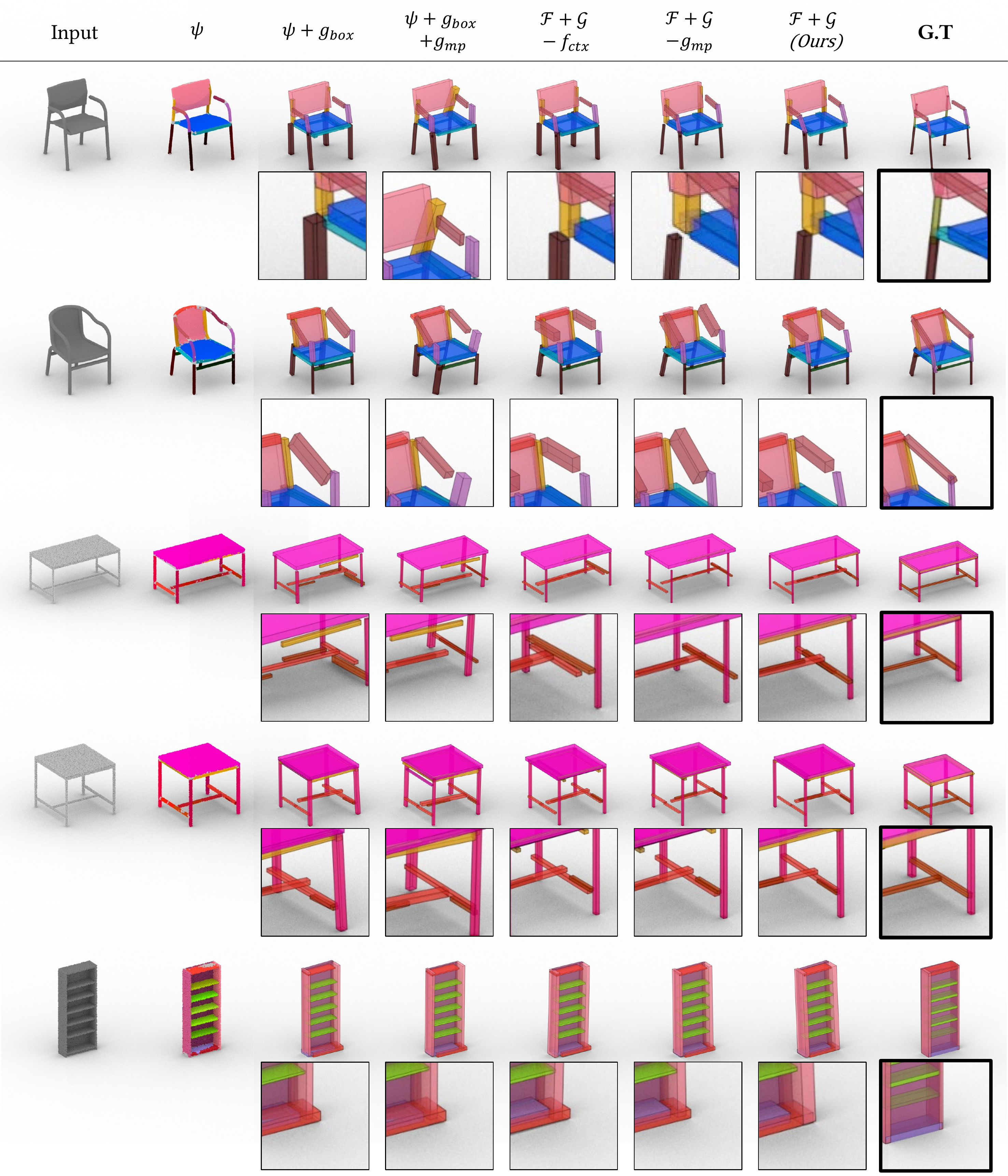}
    \caption{\textbf{The illustration of Ablation Study on Hierarchical Message Passing.} 
    The zoom-in images for each case describe the continuous improvements of the structure output more clearly. 
    As our key components for hierarchical message passing are applied, the overall arrangement of part structure improves gradually.
    \\
    \\}
    \label{fig:ablation}
\end{figure*}

\textbf{Results.}
We observe that our method based on hierarchical message passing beats the other baselines quantitatively in Table~\ref{tab:comparison}.
For the baselines with different levels of structural context, the mean part prediction accuracy also increases.
We find that learning global context even helps to improve the local part relation classification on average, which supports the necessity of our hierarchical message passing.
While the box decoder with leaf parts message passing shows compatible part prediction accuracy to ours, it extremely suffers to predict the precise part relationships with the highest edge prediction error.

We illustrate how this negatively affects to the prediction of the globally consistent structure in Figure~\ref{fig:ablation}.
As we gradually added the key component of our method, the leaf parts are coherently arranged by learning hierarchical relationships in the structure (from left to right). 
Despite the precise quality of the part segmentation output, the compared baselines fail to capture co-relations between parts across the structure.
For the box decoder baselines in the third and the fourth column, some parts pop out and degrade the overall assembly quality.
Even with part relation learning, the \emph{symmetry} between nodes is easily corrupted~(see the \emph{chair} cases).
The cases of shapes with a cluttered set of parts~(see the \emph{table} cases) get much severe where the \emph{adjacency} between parts is broken resulting in the scattered output.

For the baselines subtracting the key component of hierarchical message passing, we also found that incorporating all the context information achieves the most plausible results. 
Although the edge prediction quality seems to have a relatively small margin in numbers (Table.~\ref{tab:comparison}), we observe there is a more clear improvement in visuals.
While the symmetry between parts is preserved better than the box decoder baselines, we observe that missing one of the structural contexts still yields a flawed prediction with broken adjacency.
As each context is added, we observe the part structure accomplish \emph{globally-aligned} arrangement even with the cluttered set of parts.
For example, in the first chair case, the adjacency and alignment between a leg (brown box) and a vertical frame (yellow box) become more consistent.

\begin{table}[]
\caption{
\textbf{Comparison to Unsupervised Method.} 
Same as our paper, we use mean average precision~(mAP) with an IoU threshold 0.5. 
Note that the second column indicates which kind of supervision is used for each method.
\\
}
\resizebox{\columnwidth}{!}{
\begin{tabular}{c|c|c}
\hline
Method             & Supervision            & Chair          \\ \hline
BSP-Net~\cite{chen2020bspnet}            & None (Unsup.)           & 19.91          \\
PointGroup~\cite{jiang2020pointgroup}         & Segmentation            & 40.70          \\
\ShortName{} (\textbf{Ours}) & Segmentation, Structure & \textbf{40.81} \\ \hline
\end{tabular}
}
\label{tab:bsp-net}
\end{table}

\subsubsection{Comparison to Unsupervised Part Segmentation Method}
To demonstrate the impact of the supervision used in our framework, we compare ours with BSP-Net~\cite{chen2020bspnet}, an unsupervised method that parses 3D shapes into a set of volumetric primitives.
Since BSP-Net abstracts a raw geometry to produce \emph{super-segments}, rather than part bounding boxes, we evaluate the performance on part segmentation task only.

As BSP-Net does not predict any semantics for primitives, we assign semantics per points using ground-truth semantic labels following the original paper~\cite{chen2020bspnet} and treat predicted primitives as part instances.
For evaluation, we use mean average precision~(mAP) with an IoU threshold 0.5, as same as our paper.
In Table~\ref{tab:bsp-net}, we demonstrate the result of the quantitative evaluation for part segmentation in \emph{Chair} category.
Ours clearly outperforms the unsupervised method, while fully exploiting the supervision of part segmentation and structure both.

\subsubsection{Limitation and Discussion}
\begin{figure}
    \centering
    \includegraphics[width=\linewidth]{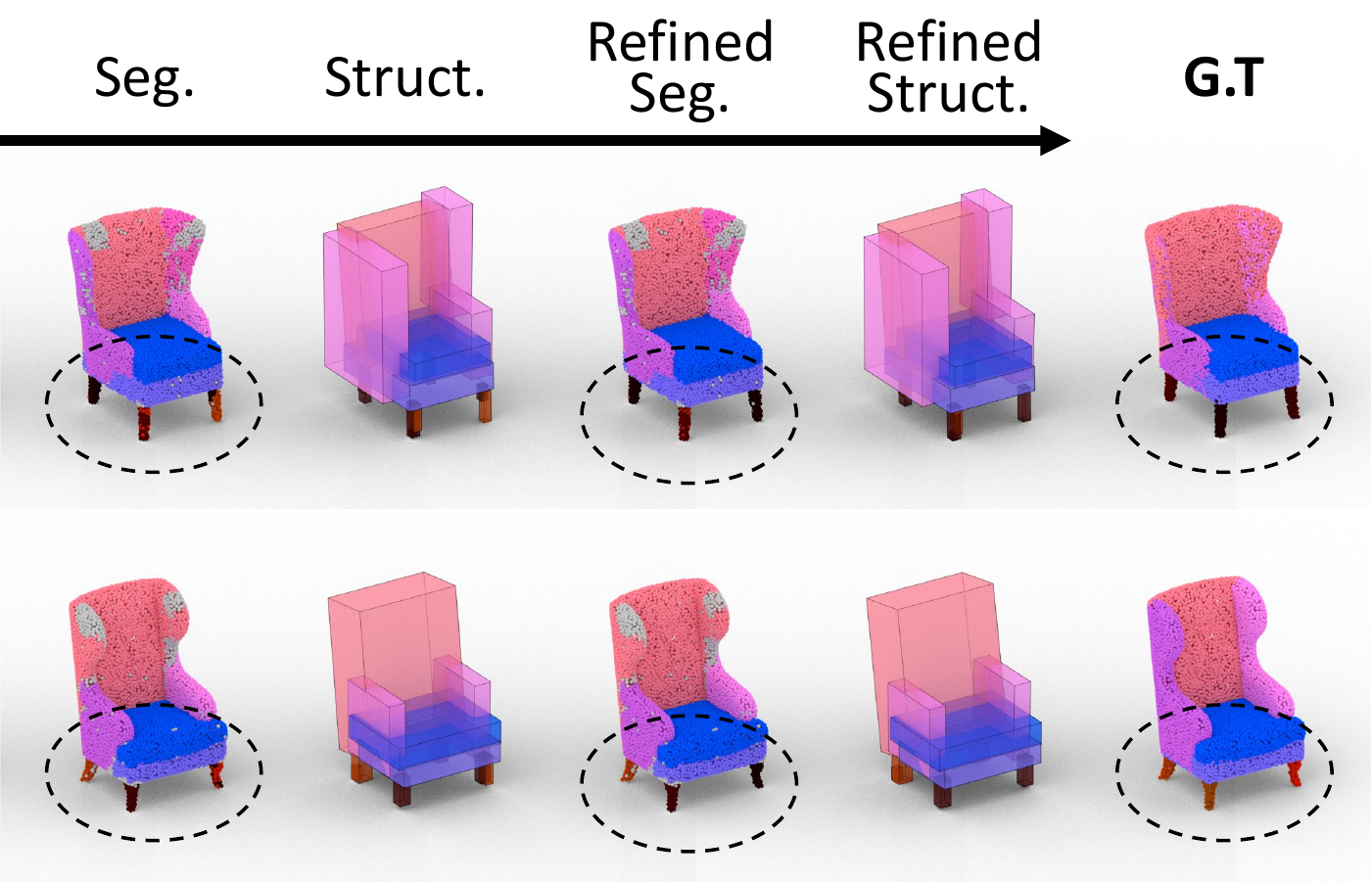}
   \caption{\textbf{Failure Cases}. Both shapes in two rows share the same Chair category. However, noisy labeling on the same region by human annotators, the orange color for \emph{foot} part and the dark brown color for \emph{leg} part, confuses the network to predict correct merge operations and yields false outputs.}
\label{fig:failure}
\end{figure}

We observe that our segmentation refinement method suffers from the imperfect supervision given by the noisy annotations.
For similar shapes, there are noisy annotations that make our network hard to predict the correct merge operation.
In Figure \ref{fig:failure}, we illustrate these failure cases caused by the noisy annotations.
For example, given almost the same shapes in \emph{Chair} category, the leg part (dark brown) is hard to be distinguished from the foot part (orange) even for the annotators.
When these confusing labels are found in the part boundaries, the network cannot clearly describe which part is falsely segmented and also examine the direction of the merge process correctly.
Unfortunately, we found these failure cases also happen across the other categories. 
Since our framework solves a supervised problem for the part segmentation, these noisy annotations largely restrain the improvement on the structure-to-segmentation refinement task and the further refinement on the part structures.

\subsubsection{More Qualitative Results} 
In this section, we provide more visuals of the qualitative evaluation on both tasks: \emph{Segmentation-to-Structure Inference} and \emph{Structure-to-Segmentation Refinement}.
As mentioned in our paper, we test our method on the three largest categories from PartNet~(e.g., chair, table, and storage furniture), and report the results for each category in the following figures. 

\begin{figure*}
    \centering
    \includegraphics{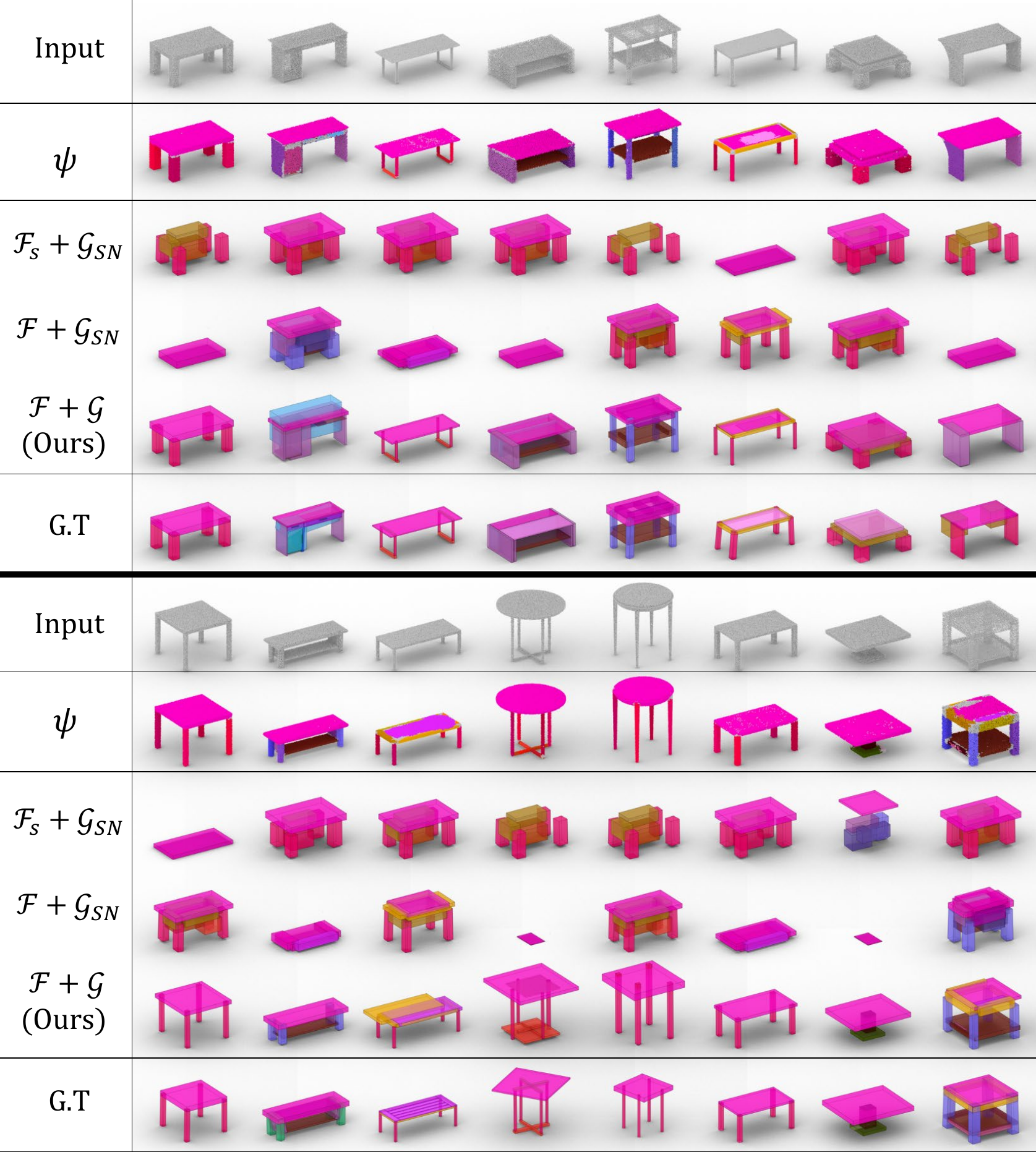}
    \caption{\textbf{More Results of Structure Inference: Table}}
    \label{fig:struct_table}
\end{figure*}

\begin{figure*}
    \centering
    \includegraphics{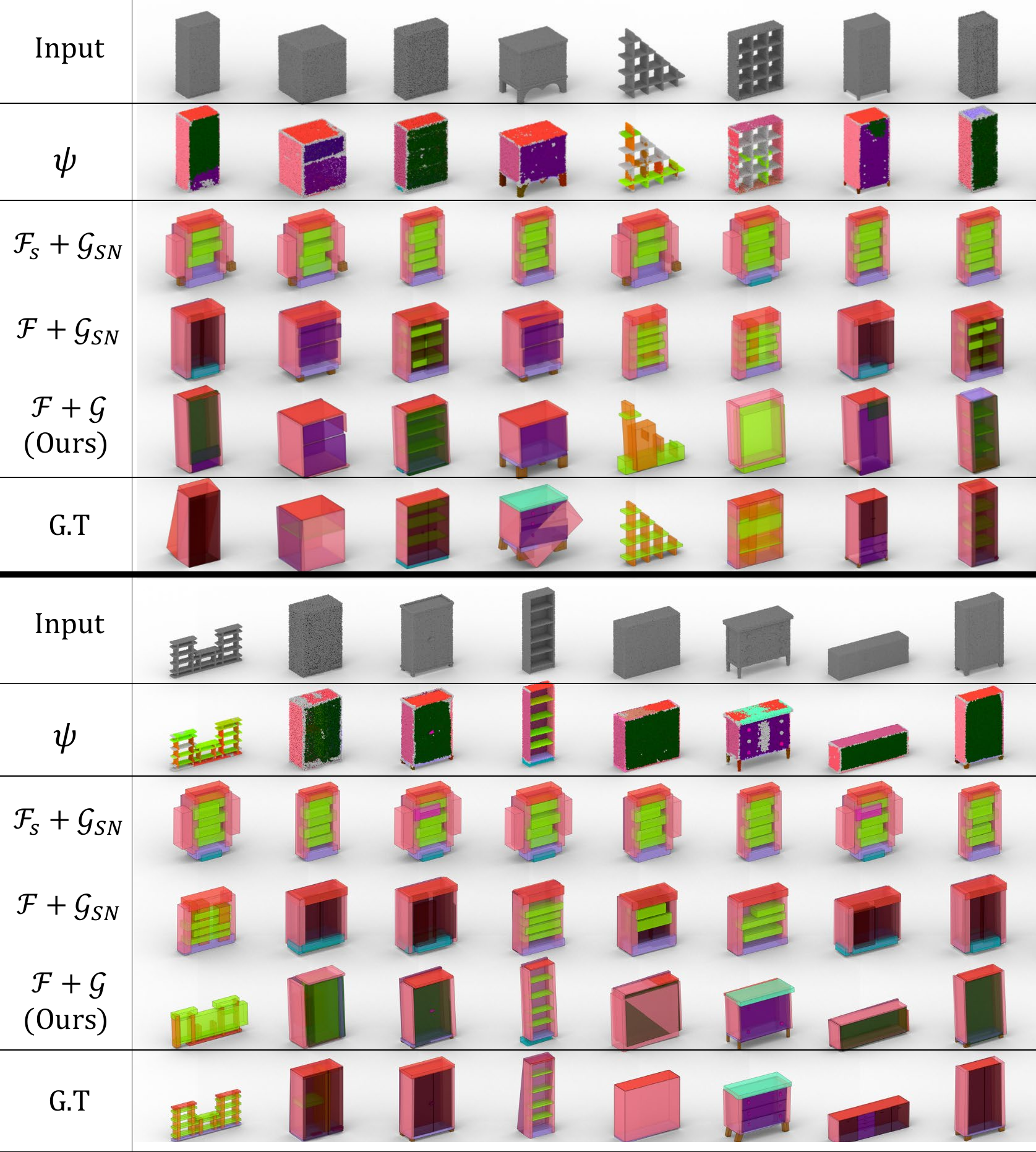}
    \caption{\textbf{More Results of Structure Inference: Storagefurniture}}
    \label{fig:struct_cabinet}
\end{figure*}

\begin{figure*}
    \centering
    \includegraphics[height=\textheight]{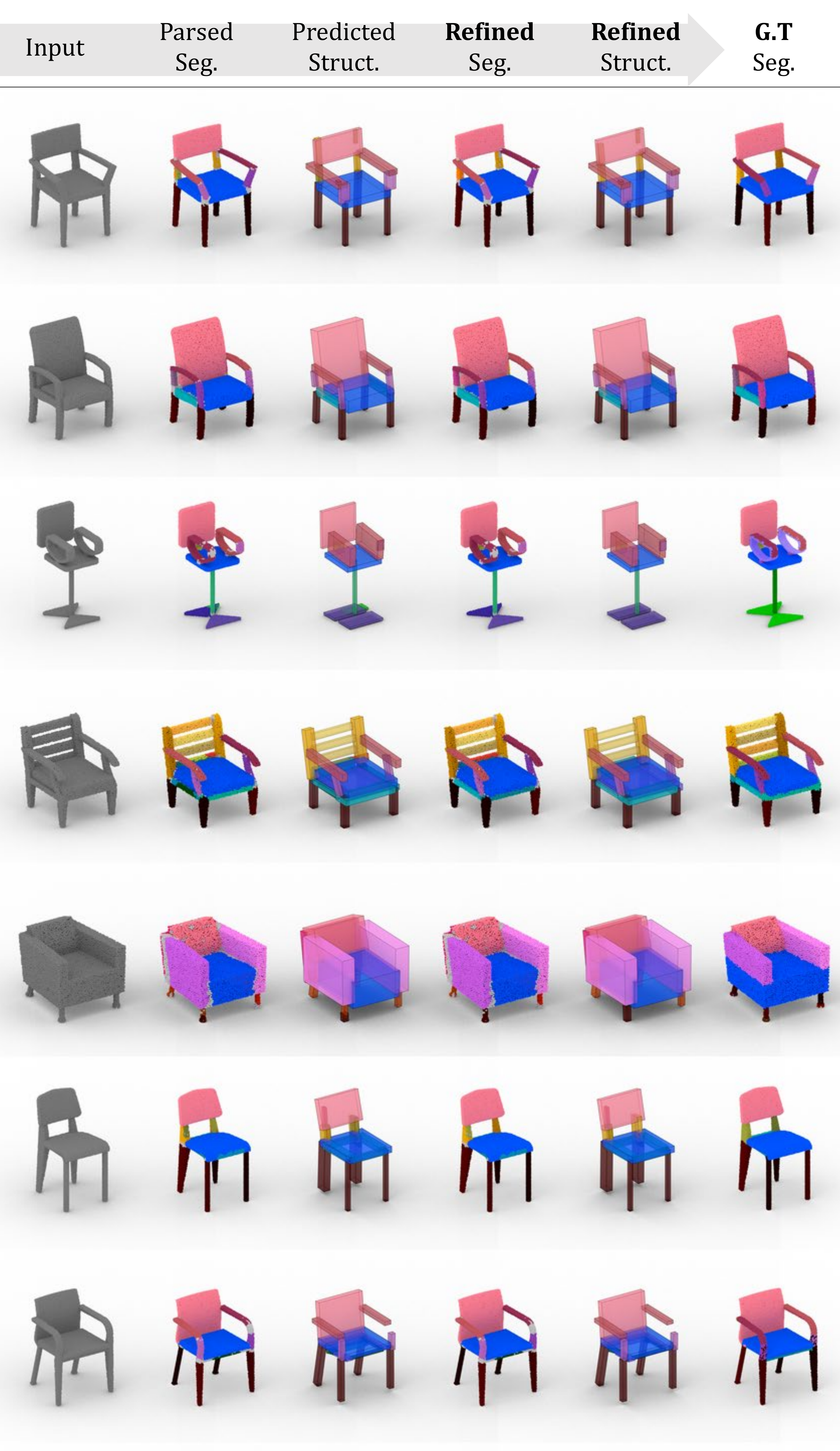}
    \caption{\textbf{More Results of The Interplay between Part Segmentation and Structure Inference: Chair}}
    \label{fig:seg_chair}
\end{figure*}

\begin{figure*}
    \centering
    \includegraphics[height=\textheight]{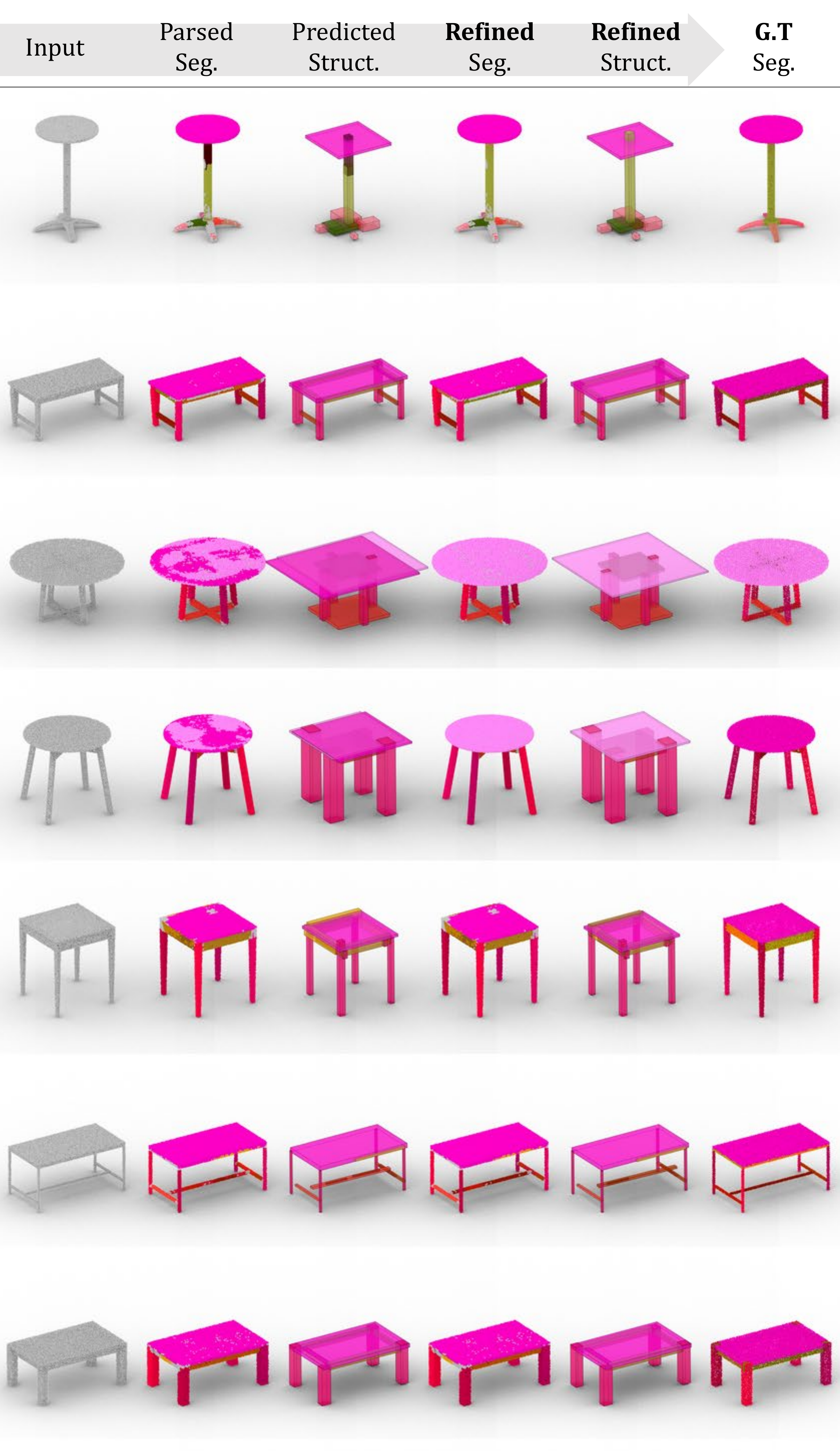}
    \caption{\textbf{More Results of The Interplay between Part Segmentation and Structure Inference: Table}}
    \label{fig:seg_table}
\end{figure*}

\begin{figure*}
    \centering
    \includegraphics[height=\textheight]{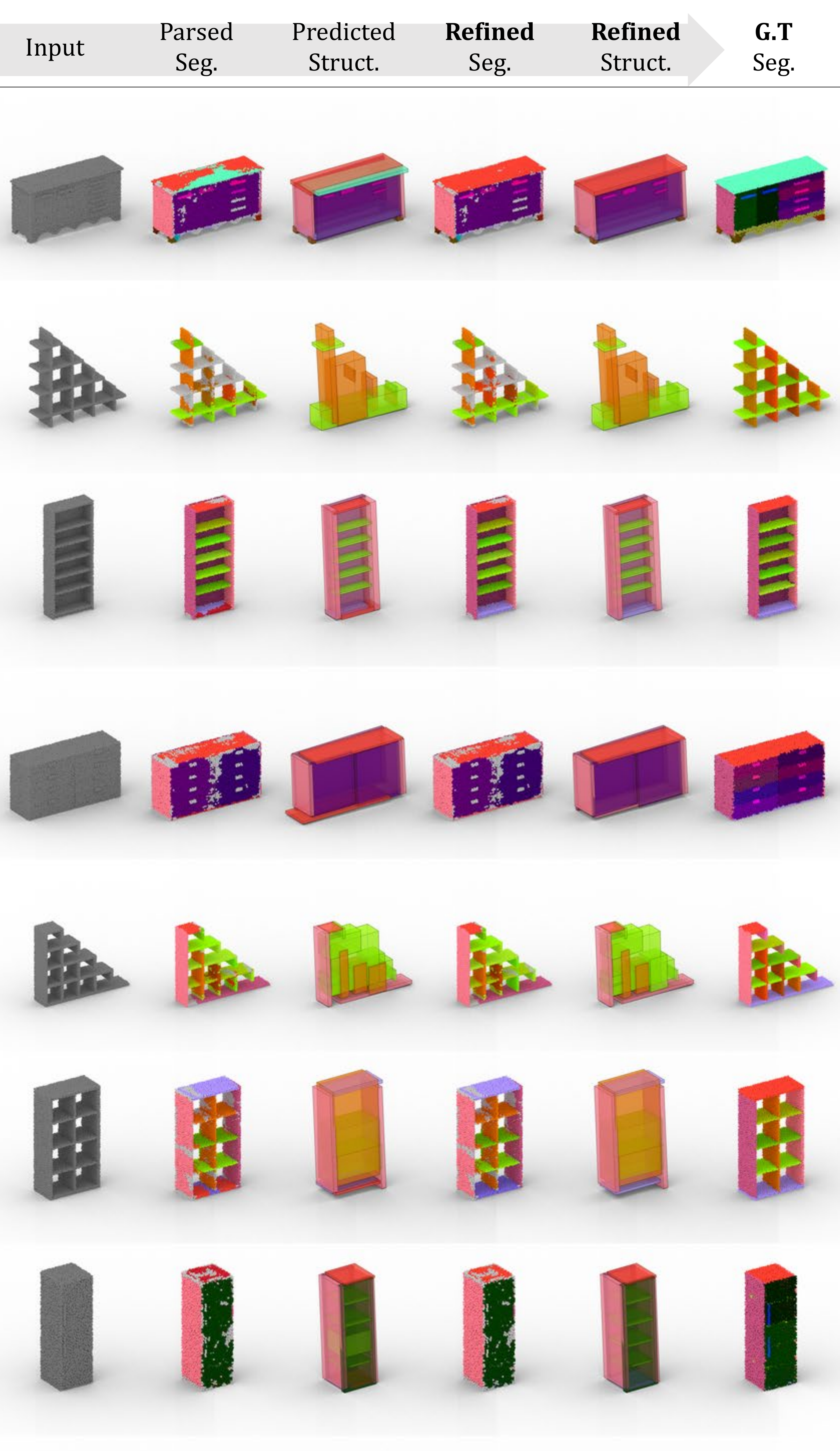}
    \caption{\textbf{More Results of The Interplay between Part Segmentation and Structure Inference: Storagefurniture}}
    \label{fig:seg_cabinet}
\end{figure*}

For the first three figures (Figure~\ref{fig:struct_chair},~\ref{fig:struct_table},~\ref{fig:struct_cabinet}), we illustrate more results of segmentation-to-structure inference network per category, comparing our method to other baselines discussed in our paper.
Next, in Figure~\ref{fig:seg_chair},~\ref{fig:seg_table},~\ref{fig:seg_cabinet}, we depict the results of structure-to-segmentation refinement including the updated structure from the refinement output. \textbf{Please see the following pages for more qualitative results.}

\clearpage

\begin{figure*}
    \centering
    \includegraphics{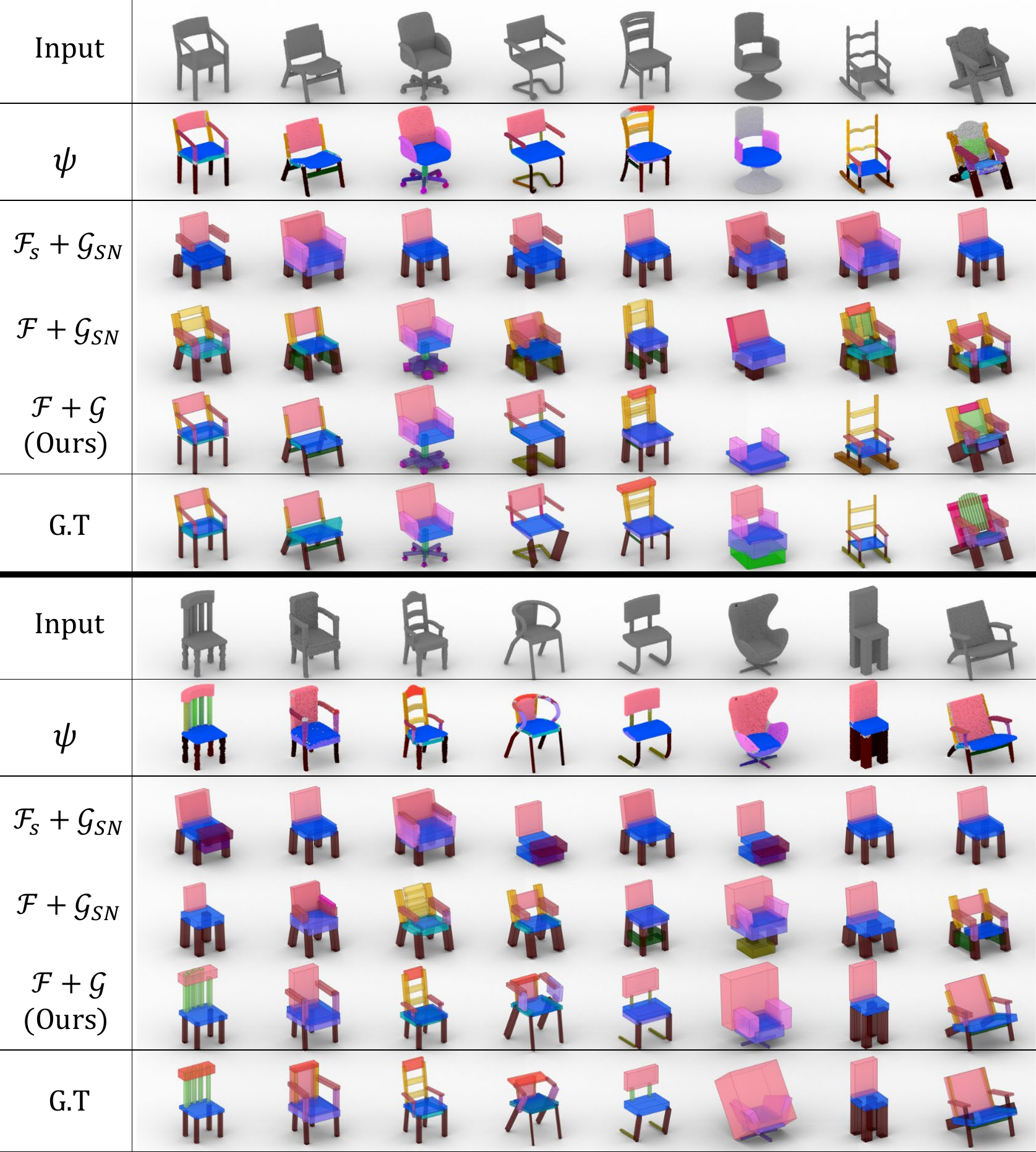}
    \caption{\textbf{More Results of Structure Inference: Chair}}
    \label{fig:struct_chair}
\end{figure*}


\end{document}